\documentclass{article}

    \PassOptionsToPackage{numbers, compress}{natbib}

    \usepackage[final]{neurips_2021}

\usepackage[utf8]{inputenc} 
\usepackage[T1]{fontenc}    
\usepackage{hyperref}       
\usepackage{url}            
\usepackage{booktabs}       
\usepackage{amsfonts}       
\usepackage{nicefrac}       
\usepackage{microtype}      
\usepackage{xcolor}         

\newif\ifupdate\updatefalse

\usepackage{mathtools}
\usepackage{stmaryrd}
\usepackage{amsfonts}
\usepackage{amsmath}
\usepackage{color}
\usepackage{bm}
\usepackage{enumitem}
\usepackage{multirow}
\usepackage{amsthm}
\usepackage{pifont}

\usepackage{subcaption}
\newcommand{\gb}[1]{\boldsymbol{#1}}
\newcommand{\taxis}{\text{ax}}
\newcommand{\targ}{\text{ap}}

\newcommand{\ee}{\mathbf{v}}
\newcommand{\qe}{\mathbf{V}_q}

\newcommand{\ttrain}{\text{train}}
\newcommand{\tvalid}{\text{valid}}
\newcommand{\ttest}{\text{test}}
\newcommand{\cl}{\textbf{cl}}

\newcommand{\betae}{{\normalsize{B}\footnotesize{ETA}\normalsize{E}}}

\newtheorem{prop}{Proposition}
\newtheorem{lemma}{Lemma}
\newtheorem{definition}{Definition}

\newcommand{\udfsection}[1]{\noindent\textbf{#1}\, }

\newcommand{\std}[1]{\tiny{$\pm$#1}}
\renewcommand{\subsubsection}[1]{\textbf{#1:}}

\title{ConE: Cone Embeddings for Multi-Hop Reasoning over Knowledge Graphs}

\author{%
  Zhanqiu Zhang$^{1,2}$ \qquad
  Jie Wang$^{1,2}$ \thanks{Corresponding author.} \qquad
  Jiajun Chen$^{1,2}$ \qquad
  Shuiwang Ji$^3$ \qquad
  Feng Wu$^{1,2}$ \qquad
  \vspace{1mm}\\
$^1$CAS Key Laboratory of Technology in GIPAS\\
University of Science and Technology of China\\
$^2$Institute of Artificial Intelligence\\
Hefei Comprehensive National Science Center\\
\texttt{\{zqzhang,jjchen\}@mail.ustc.edu.cn,\{jiewangx,fengwu\}@ustc.edu.cn}\\
$^3$Texas A\&M University\\
\texttt{sji@tamu.edu}
}

\begin{document}
\maketitle

\begin{abstract}
Query embedding (QE)---which aims to embed entities and first-order logical (FOL) queries in low-dimensional spaces---has shown great power in multi-hop reasoning over knowledge graphs. 
Recently, embedding entities and queries with geometric shapes becomes a promising direction, as geometric shapes can naturally represent answer sets of queries and logical relationships among them.
However, existing geometry-based models have difficulty in modeling queries with negation, which significantly limits their applicability. 
To address this challenge, we propose a novel query embedding model, namely \textbf{Con}e \textbf{E}mbeddings (ConE), which is the first geometry-based QE model that can handle all the FOL operations, including conjunction, disjunction, and negation.
Specifically, ConE represents entities and queries as Cartesian products of two-dimensional cones, where the intersection and union of cones naturally model the conjunction and disjunction operations. By further noticing that the closure of complement of cones remains cones, we design geometric complement operators in the embedding space for the negation operations.
Experiments demonstrate that ConE significantly outperforms existing state-of-the-art methods on benchmark datasets.
\end{abstract}

\section{Introduction}\label{sec:intro}

Multi-hop reasoning over knowledge graphs (KGs)---which aims to find answer entities of given queries using knowledge from KGs---has attracted great attention from both academia and industry recently \cite{pullnet,emb_kgqa,alime}. In general, it involves answering first-order logic (FOL) queries over KGs using operators including existential quantification ($\exists$), conjunction ($\land$), disjunction ($\lor$), and negation ($\neg$). 
A popular approach to multi-hop reasoning over KGs is to first transform a FOL query to its corresponding computation graph---where each node represents a set of entities and each edge represents a logical operation---and then traverse the KG according to the computation graph to identify the answer set.
However, this approach confronts two major challenges. First, when some links are missing in KGs, it has difficulties in identifying the correct answers. Second, it needs to deal with all the intermediate entities on reasoning paths, which may lead to exponential computation cost.

To address these challenges, researchers have paid increasing attention to the query embedding (QE) technique, which embeds entities and FOL queries in low-dimensional spaces \cite{gqe,q2b,beta,faith}.  
QE models associate each logical operator in computation graphs with a logical operation in embedding spaces.
Given a query, QE models generate query embeddings following the corresponding computation graph.
Then, they determine whether an entity is a correct answer based on similarities between the query embeddings and entity embeddings. 

Among the existing QE models, geometry-based models that embed entities and queries into geometric shapes have shown promising performance \cite{gqe,traversing,chains,q2b}. Geometry-based models usually represent entity sets as "regions" (e.g., points and boxes) in Euclidean spaces and then design set operations upon them. For example, Query2Box \cite{q2b} represents entities as points and queries as boxes. If a point is inside a box, then the corresponding entity is the answer to the query. Compared with non-geometric methods, geometric shapes provide a natural and easily interpretable way to represent sets and logical relationships among them.

However, existing geometry-based models have difficulty in modeling queries with negations, which significantly limits their applicability. For example, GQE \cite{gqe} and Query2Box \cite{q2b}---which embed queries to points and boxes, respectively---cannot handle queries with negation, as the complement of a point/box is no longer a point/box. To tackle this problem, \citet{beta} propose a probabilistic QE model using Beta distributions. However, it does not have some advantages of geometric models. For example, using Beta distributions, it is unclear how to determine whether an entity is an answer to a query as that in the box case \cite{q2b}. Therefore, proposing a geometric QE model that can model all the FOL queries is still challenging but promising.

In this paper, we propose a novel geometry-based query embedding model---namely, \textbf{Con}e \textbf{E}mbeddings (ConE)---which represents entities and queries as Cartesian products of two-dimensional cones. Specifically, if the cones representing entities are subsets of the cones representing queries, then these entities are the answers to the query.
To perform multi-hop reasoning in the embedding space, we define the conjunction and disjunction operations that correspond to the intersection and union of cones.
Further, by noticing that the closure of complement of cones are still cones, we correspondingly design geometric complement operators in the embedding space for the negation operations.
To the best of our knowledge, ConE is the first geometry-based QE model that can handle all the FOL operations, including conjunction, disjunction, and negation.
Experiments demonstrate that ConE significantly outperforms existing state-of-the-art methods on benchmark datasets.

\section{Related Work}
Our work is related to answering multi-hop logical queries over KGs and geometric embeddings.

\textbf{Answering multi-hop logical queries over KGs.} To answer multi-hop FOL queries, path-based methods \cite{deeppath,rl,ruge} start from anchor entities and require traversing the intermediate entities on the path, which leads to exponential computation cost. Embedding-based models are another line of works, which embed FOL queries into low-dimensional spaces. For example, existing works embed queries to geometric shapes \cite{gqe,q2b,traversing,chains}, probability distributions \cite{beta}, and complex objects \cite{quantum,faith}. Our work also embeds queries to geometric shapes. The main difference is that our work can handle all the FOL operations, while existing works cannot.

\textbf{Other Geometric embeddings.} Geometric embeddings are popular in recent years. For example, geometric operations including translation \cite{transe,transr}, rotation \cite{complex,rotate,n3,dura}, and complex geometric operations \cite{poincare_kg,hake} have been widely used in knowledge graph embeddings. Other geometric embedding methods also manage to use boxes \cite{improve_box}, convex cones \cite{denotion_prob,order_emb,cone_ijcai}, etc. For example, \citet{cone_ijcai} use axis-aligned cones to embed ontologies expressed in the ALC description logic, and use polars of cones to model negation operators.   
Recent years have also witnessed the development of embeddings in non-Euclidean geometry, such as Poincar\'e embeddings \cite{poincare_kg,poincare} and hyperbolic entailment cones \cite{ent_cone}. 
Notably, although there exist works that also use cone embeddings \cite{denotion_prob,order_emb,ent_cone,cone_ijcai}, they are not designed for the multi-hop reasoning task and their definition of cones are different from that in our work.

\section{Preliminaries}\label{sec:preliminaries}
In this section, we review the background of query embeddings in Section \ref{sec:background} and introduce some basic concepts of two-dimensional cones in Section \ref{sec:cones}.
\subsection{Backgrounds}\label{sec:background}

\textbf{Knowledge Graphs (KGs). } Given a set $\mathcal{V}$ of entities (vertices) and a set $\mathcal{E}$ of relations (edges), a knowledge graph $\mathcal{G}=\{(s_i, p_j, o_k)\}\subset\mathcal{V}\times \mathcal{E}\times\mathcal{V}$ is a set of factual triple, where $p_j\in\mathcal{E}$ is a predicate, and $s_i,o_k\in\mathcal{V}$ are subject and object, respectively. Suppose that $r_j(\cdot,\cdot) \in\mathcal{R}$ is a binary function $r_j:\mathcal{V}\times \mathcal{V}\rightarrow\{\text{True}, \text{False}\}$ corresponding to $p_j$, where $r_j(s_i,o_k)=\text{True}$ if and only if $(s_i, p_j, o_k)$ is a factual triples. Then, for all $(s_i, p_j, o_k)\in\mathcal{G}$, we have $r_j(s_i,o_k)=\text{True}$. Note that both $\mathcal{E}$ and $\mathcal{R}$ are involved with relations, while $\mathcal{E}$ is a set of relation instances and $\mathcal{R}$ is a set of relational functions.

\textbf{First-Order Logic (FOL). } FOL queries in the query embedding literature involve logical operations including existential quantification ($\exists$), conjunction ($\land$), disjunction ($\lor$), and negation ($\neg$). Universal quantification ($\forall$) is not included, as no entity connects with all other entities in real-world KGs \cite{beta}.

\begin{figure}[t]
\begin{subfigure}{0.5\textwidth}
  \includegraphics[width=200pt]{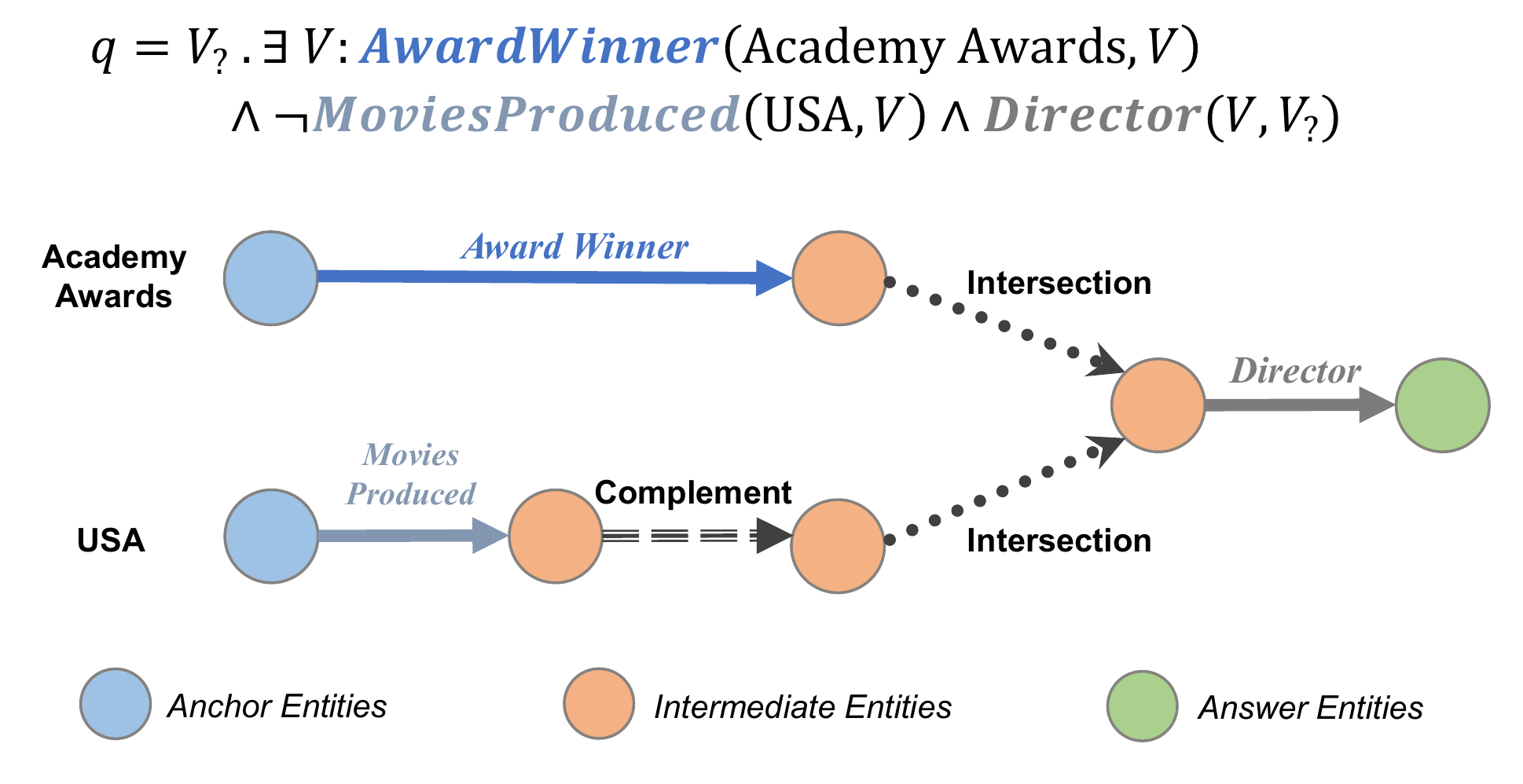}
\caption{A Computation Graph}\label{subfig:comp_graph}
\end{subfigure}\hspace{18mm}
\begin{subfigure}{0.3\textwidth}
  \includegraphics[width=130pt]{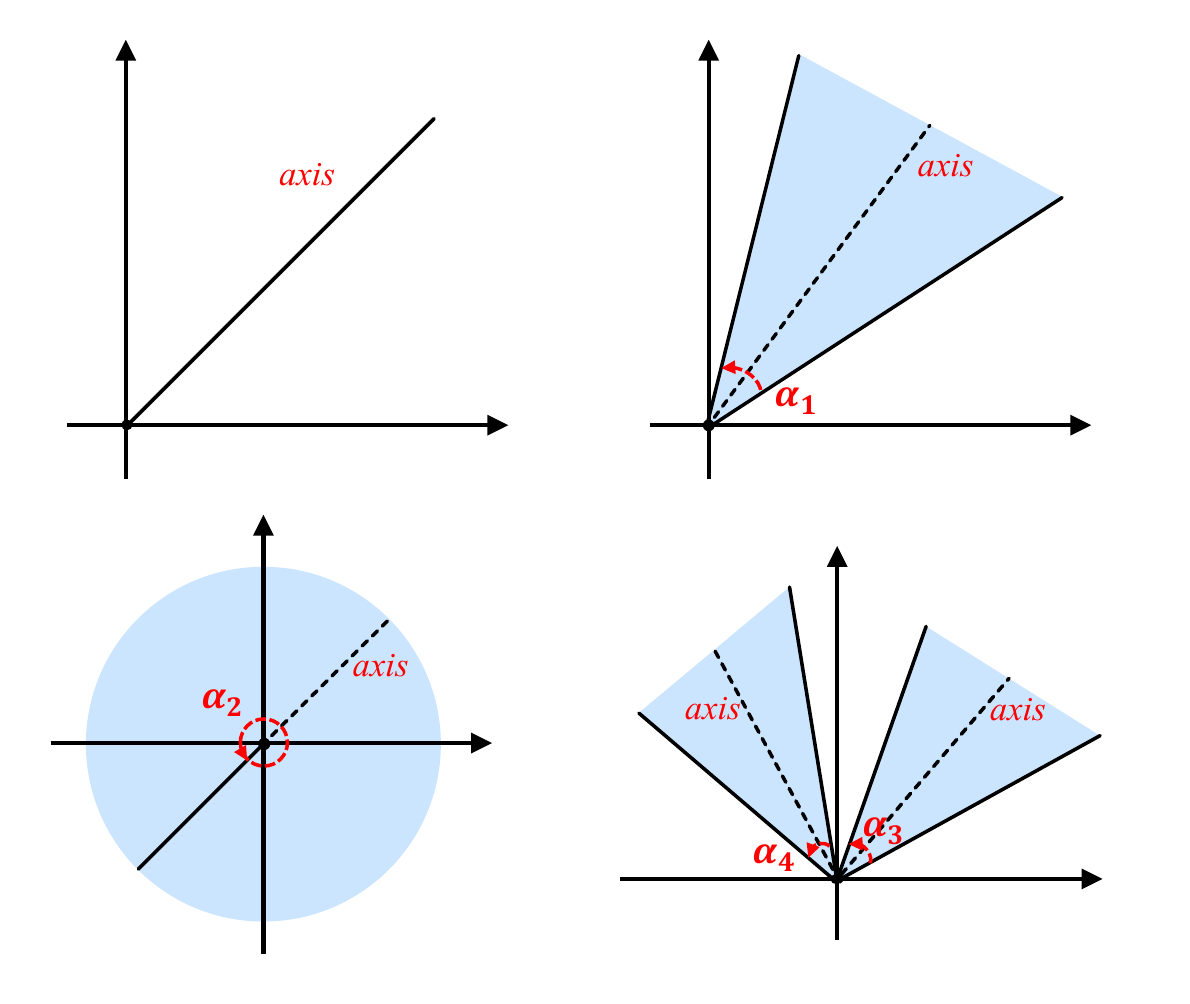}
    \vspace{-6mm}
  \caption{Cones}\label{subfig:cones}
\end{subfigure}\hfil 
\caption{Examples of a computation graph and several cones. In Figure \ref{subfig:comp_graph}, the natural language interpretation of the query is "List all the directors of non-American movies that win the Academy Awards". In Figure \ref{subfig:cones}, the four cones are: a cone with aperture $0$, a sector-cone with aperture $0<\alpha_1<\pi$, a sector-cone with aperture $\alpha_2=2\pi$, and a cone that is the union of two sector-cones.}
\vspace{-3mm}
\end{figure}

We use FOL queries in its Disjunctive Normal Form (DNF) \cite{dnf}, which represents FOL queries as a disjunction of conjunctions.
To formulate FOL queries, we assume that $\mathcal{V}_a\subset \mathcal{V}$ is the non-variable anchor entity set, $V_1,\dots,V_k$ are existentially quantified bound variables, and $V_?$ is the target variable, i.e., the answers to a certain query. Then, a FOL query $q$ in the disjunctive normal form is:
\begin{align*}
    q[V_?]=V_?.\ \exists\,V_1,\ldots,V_k:c_1\lor c_2\lor \dots \lor c_n.
\end{align*}
Specifically, $c_i$ are \textit{conjunctions}, i.e., $c_i=e_{i1}\land \cdots\land e_{im}$, where $e_{ij}=r(v_a,V)$ or $\neg r(v_a,V)$ or $r(V',V)$ or $\neg r(V',V)$, $v_a\in\mathcal{V}_a,  V\in\{V_?,V_1,\dots,V_k\}$, $V'\in\{V_1,\dots,V_k\}$, and $ V\ne V'.$

Using the aforementioned notations, answering a query $q$ is equivalent to \textit{finding the set of entities} $\llbracket q\rrbracket\subset \mathcal{V}$, where $v\in\llbracket q\rrbracket$ if and only if $q[v]$ is True. 

\textbf{Computation Graphs. } Given a query, we represent the reasoning procedure as a computation graph (see Figure \ref{subfig:comp_graph} for an example),
of which nodes represent entity sets and edges represent logical operations over entity sets. 
We map edges to logical operators according to the following rules. 
\vspace{-2mm}
\begin{itemize}[leftmargin=8mm]
    \item \textit{Relation Traversal}$\rightarrow$\textit{Projection Operator} $\mathcal{P}$. Given a set of entities $S\subset \mathcal{V}$ and a relational function $r\in\mathcal{R}$, the projection operator $\mathcal{P}$ outputs all the adjacent entities $\cup_{v\in S}N(v,r)$, where $N(v,r)$ is the set of entities such that $r(v,v')=\text{True}$ for all $v'\in N(v,r)$.
    \item \textit{Conjunction}$\rightarrow$\textit{Intersection Operator} $\mathcal{I}$. Given $n$ sets of entities $\{S_1,S_2,\dots,S_n\}$, the intersection operator $\mathcal{I}$ performs set intersection to obtain $\cap_{i=1}^n S_n$.
    \item \textit{Disjunction}$\rightarrow$\textit{Union Operator} $\mathcal{U}$. Given $n$ sets of entities $\{S_1,S_2,\dots,S_n\}$, the union operator $\mathcal{U}$ performs set union to obtain $\cup_{i=1}^n S_n$.
    \item \textit{Negation}$\rightarrow$\textit{Complement Operator} $\mathcal{C}$. Given an entity set $S\subset \mathcal{V}$, $\mathcal{C}$ gives $\bar{S}=\mathcal{V}\backslash S$.
\end{itemize}

\textbf{Query Embeddings (QE). } QE models generate low-dimensional continuous embeddings for queries and entities, and associate each logical operator for entity sets with an operation in embedding spaces. 
Since an entity is equivalent to a set with a single element and each query $q$ is corresponding to a unique answer set $\llbracket q\rrbracket$, the aim of QE models is equivalent to embedding entity sets that can be answers to some queries.

\subsection{Cones in Two-Dimensional Spaces}\label{sec:cones}
To represent FOL queries as Cartesian products of two-dimensional cones, we introduce some definitions about cones and the parameterization method of a special class of cones.
\begin{definition}[\citet{boyd}]\label{def:cone}
A set $C\subset\mathbb{R}^2$ is called a \textbf{cone}, if for every $x\in C$ and $\lambda\ge 0$, we have $\lambda x\in C$ . A set is a \textbf{convex cone} if it is convex and a cone, which means that for any $x_1, x_2\in C$ and $\lambda_1, \lambda_2\ge 0$, we have $\lambda_1 x_1+\lambda_2 x_2\in C$.
\end{definition}

By letting $\lambda=0$ in Definition \ref{def:cone}, we know that a cone must contain the origin. In view of this property, we define a new operation called \textit{closure-complement} for cones.
\begin{definition}\label{def:odc}
Suppose that $C\subset\mathbb{R}^2$ is a cone. Then, the \textbf{closure-complement} of $C$ is defined by $\tilde{C}=\cl(\mathbb{R}^2\backslash C)$, where $\cl(\cdot)$ is the closure of a set.
\end{definition}

Next, we introduce a class of cones that can be parameterized in a scalable way.
\begin{definition}\label{def:sec_cone}
A 2D closed cone is called a \textbf{sector-cone}, if its closure-complement or itself is convex.
\end{definition}
The set of sector-cones is closed under closure-complement and their union and intersection are still cones. Besides, we have the following proposition, whose proof is provided in Appendix A.
\begin{prop}\label{prop:axial}
A sector-cone is always axially symmetric.
\end{prop}

\textbf{Parameterization of 2D Sector-Cones.}
Proposition \ref{prop:axial} suggests that we can use a pair of parameters to represent a two-dimensional sector-cone:
\begin{align*}
    (\theta_{\taxis}, \theta_{\targ}), \text{where }  \theta_{\taxis}\in [-\pi,\pi),\ \theta_{\targ}\in [0,2\pi].
\end{align*}
Specifically, $\theta_{\taxis}$ represents the angle between the symmetry axis of the sector-cone and the positive $x$ axis. $\theta_{\targ}$ represents the aperture of the sector-cone. For any points in the cone, its phase will be in $[\theta_{\taxis}-\theta_{\targ}/2, \theta_{\taxis}+\theta_{\targ}/2]$.
Figure \ref{subfig:cones} gives examples of several (sector-)cones. One may notice that sector-cones share some similarities with boxes defined in Query2Box, which also involves region representations. However, we argue that sector-cones are more expressive than boxes, of which the details are provided in Appendix F.

Let $\mathbb{K}$ be the space consisting of all $(\theta_{\taxis}, \theta_{\targ})$. We can represent an arbitrary sector-cone $C_0$ as $C_0=(\theta_{\taxis}, \theta_{\targ})\in \mathbb{K}$.
Then, for a $d$-ary Cartesian product of sector-cones
\begin{align}\label{eqn:cartesian}
     C=C_1\times C_2\times \ldots\times C_d,
\end{align}
we represent it via a $d$-dimensional vector in $\mathbb{K}^d$:
\begin{align}\label{eqn:param}
    C=\left((\theta_{\taxis}^1, \theta_{\targ}^1), \dots, (\theta_{\taxis}^d, \theta_{\targ}^d)\right)\subset \mathbb{K}^d.
\end{align}
where $\theta_{\taxis}^i\in [-\pi,\pi),\ \theta_{\targ}^i\in [0,2\pi]$, for $i=1,\dots,d$.
Or equivalently, $C=(\gb{\theta}_{\taxis}, \gb{\theta}_{\targ})$,
where $\gb{\theta}_{\taxis}=(\theta_{\taxis}^1,\dots,\theta_{\taxis}^d)\in[-\pi,\pi)^d$ and $\gb{\theta}_{\targ}=(\theta_{\targ}^1,\dots,\theta_{\targ}^d)\in[0,2\pi]^d$.

\section{Cone Embeddings}\label{sec:methods}
In this section, we propose \textbf{Con}e \textbf{E}mbeddings (ConE) for multi-hop reasoning over KGs.
We first introduce cone embeddings for conjunctive queries and entities in Section \ref{sec:cone_e}. Afterwards, we introduce the logical operators and the methods to learn ConE in Sections \ref{sec:logical_op} and \ref{sec:train_obj}. 

\subsection{Cone Embeddings for Conjunctive Queries and Entities}\label{sec:cone_e}
As introduced in Section \ref{sec:background}, conjunctive queries constitute the basis of all queries in the DNF form. Embeddings of all queries can be generated by applying logical operators to conjunctive queries' embeddings. Thus, we design embeddings for conjunctive queries in this section. We model queries with disjunction using the Union Operator $\mathcal{U}$ in Section \ref{sec:logical_op}.

In general, the answer entities to a conjunctive query $q$ have similar semantics. For example, answers to the query that "List all the directors of American movies" should all be persons; answers to the query that "List all the Asian cities that ever held Olympic Games" should all be places. If we embed an entity set $\llbracket q\rrbracket$ into an embedding space, we expect entities in $\llbracket q\rrbracket$ to have similar embeddings. Thus, we expect their embeddings to form a "region" in the embedding space. If the embedding of an entity is inside the region, then the entity is likely to be an answer.  Further, we can find a semantic center and a boundary for the region, where the semantic center represents the semantics of $\llbracket q\rrbracket$ and the boundary designates how many entities are in $\llbracket q\rrbracket$.

To model the embedding region of $\llbracket q\rrbracket$, we propose to embed it to a Cartesian product of \textit{sector-cones}. Specifically, we use the parameter $\theta_{\taxis}^i$ to represent the semantic center, and the parameter $\theta_{\targ}^i$ to determine the boundary of $\llbracket q\rrbracket$. If we use a $d$-ary Cartesian product, i.e., the embedding dimension is $d$, we define the embedding of $\llbracket q\rrbracket$ as
\begin{align*}
    \textbf{V}_q^c=(\gb{\theta}_{\taxis}, \gb{\theta}_{\targ}),
\end{align*}
where $\gb{\theta}_{\taxis}\in[-\pi,\pi)^d$ are \textit{axes} and $\gb{\theta}_{\targ}\in[0,2\pi]^d$ are \textit{apertures}.

An entity $v\in\mathcal{V}$ is equivalent to an entity set with a single element, i.e., $\{v\}$. We propose to represent an entity as a Cartesian product of cones with apertures $0$, where the axes indicates the semantics of the entity. Formally, if the embedding dimension is $d$, the cone embedding of $v$ is $\textbf{v}=(\gb{\theta}_{\taxis},\textbf{0})$, where $\gb{\theta}_{\taxis}\in[-\pi,\pi)^d$ is the axis embedding and $\textbf{0}$ is a $d$-dimensional vector with all elements being $0$.

\begin{figure}[t]
\vspace{-1mm}
\centering 
\begin{subfigure}{0.2\columnwidth}
  \includegraphics[width=85pt]{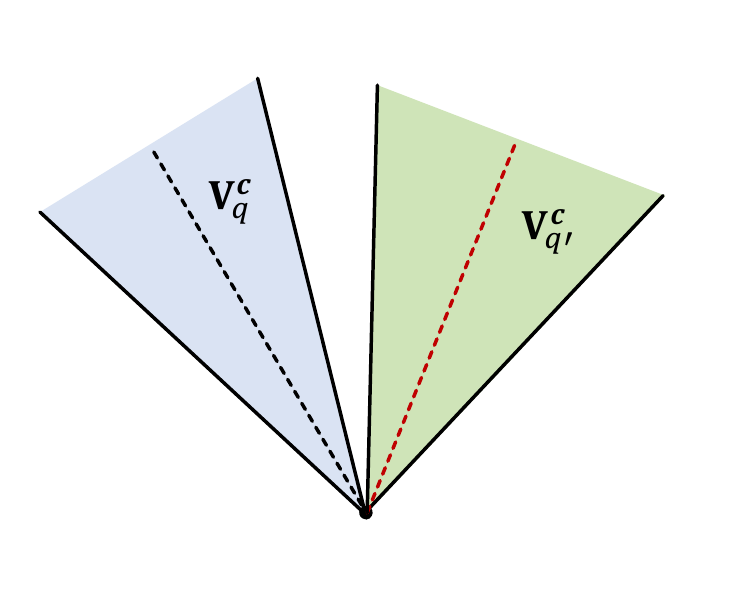}
  \vspace{-5mm}
\caption{Projection}\label{subfig:projection}
\end{subfigure}\hfil 
\begin{subfigure}{0.2\columnwidth}
  \includegraphics[width=85pt]{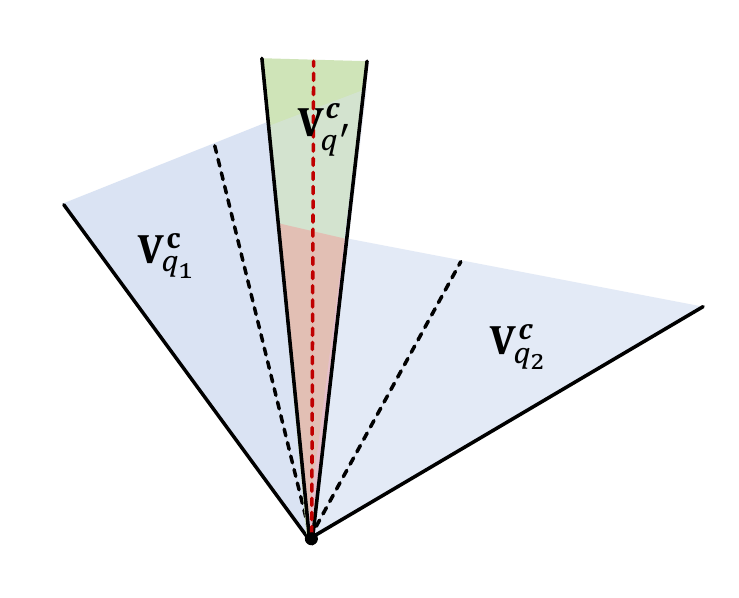}
    \vspace{-5mm}
  \caption{Intersection}\label{subfig:intersection}
\end{subfigure}\hfil 
\begin{subfigure}{0.2\columnwidth}
  \includegraphics[width=85pt]{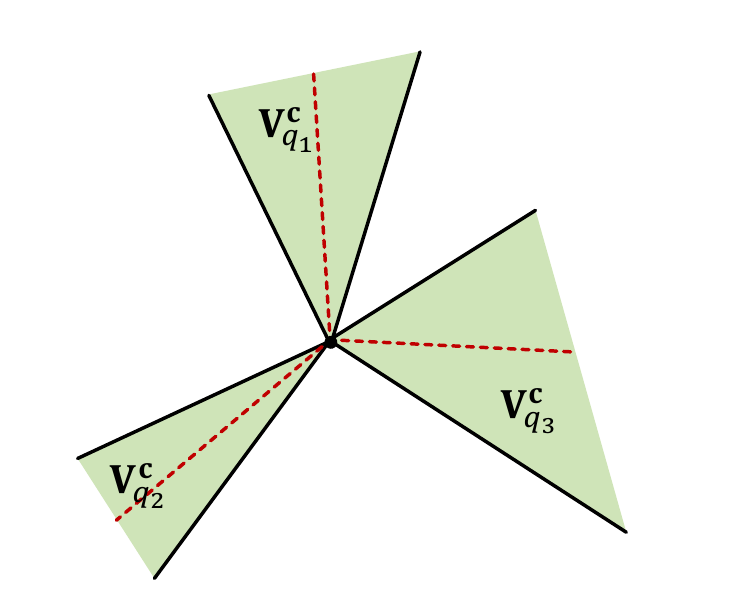}
    \vspace{-5mm}
  \caption{Union}\label{subfig:union}
\end{subfigure}\hfil 
\begin{subfigure}{0.2\columnwidth}
  \includegraphics[width=85pt]{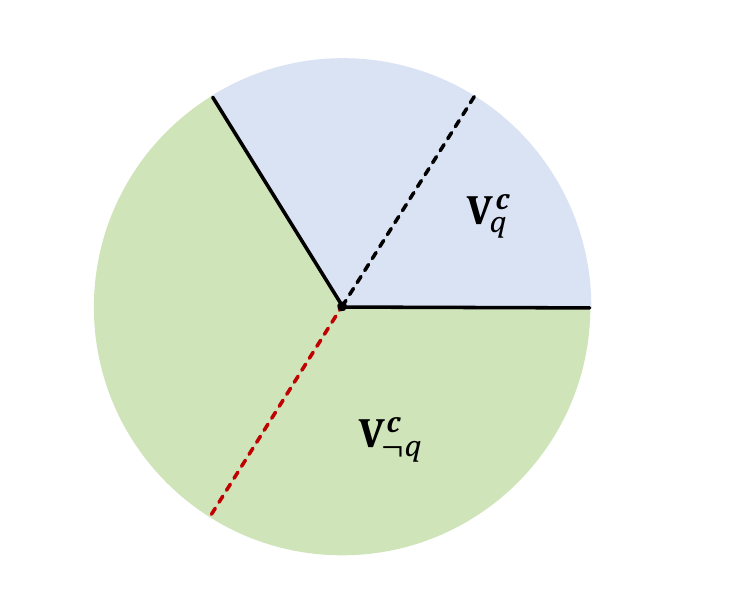}
    \vspace{-5mm}
  \caption{Complement}\label{subfig:complement}
\end{subfigure}\hfil 
\begin{subfigure}{0.2\columnwidth}
  \includegraphics[width=85pt]{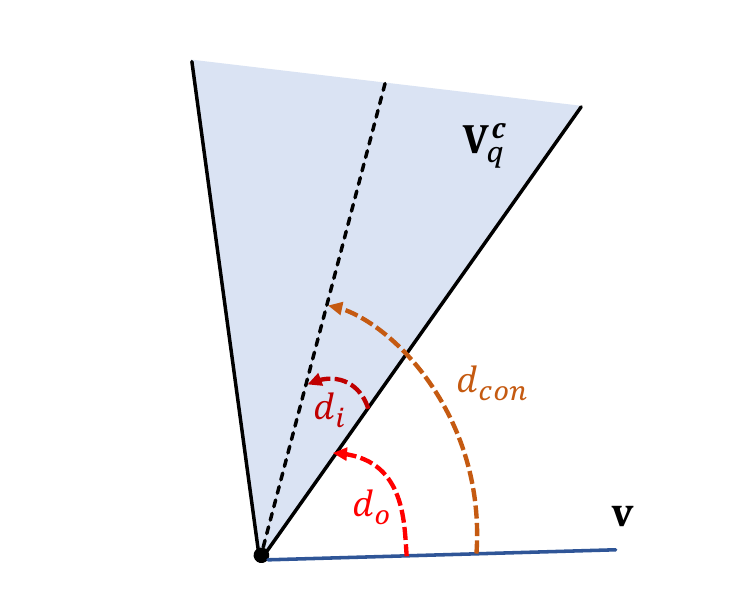}
    \vspace{-5mm}
  \caption{Distance}\label{subfig:distance}
\end{subfigure}\hfil 
\vskip -1mm
\caption{ConE's logical operators and distance function, of which the embedding dimension $d=1$.}
\label{fig:models}
\vspace{-4mm}
\end{figure}

\subsection{Logical Operators for Cone Embeddings}\label{sec:logical_op}
\vspace{-1mm}
In this section, we introduce our designed logical operators of ConE in the embedding space, including projection, intersection, union, and complement.

It is worth noting that, the composition of logical operators may lead to non-sense queries. For example, the queries "List the intersection/union of American movies and non-American movies" and "List the intersection of American movies and Asian movies" make no sense in real-world applications. However, the main aim of a query embedding model is to represent all entity sets that \textit{can be answer to some real-world query}. Therefore, we do not need to model the entity sets that only correspond to theoretically possible queries \cite{beta}.

\textbf{Projection Operator} $\mathcal{P}$\textbf{.} The goal of $\mathcal{P}$ is to represent an entity's adjacent entities that are linked by a given relation. 
It maps an entity set to another entity set (see Figure \ref{subfig:projection}). Thus, we define a relation-dependent function in the embedding space for $\mathcal{P}$:
\begin{align*}
    f_r:\mathbb{K}^d\rightarrow\mathbb{K}^d,\ \textbf{V}_q^c\mapsto \textbf{V}_q^{c'}.
\end{align*}
We implement $f_r$ in a neural way. First, we represent relations as relational translations of query embeddings and assign each relation with an embedding $\textbf{r}=(\gb{\theta}_{\taxis,r}, \gb{\theta}_{\targ,r})$. Then, we define $f_r$ as 
\begin{align}\label{eqn:proj_op}
    f_r(\textbf{V}_q)&=g(\textbf{MLP}([\gb{\theta}_{\taxis}+\gb{\theta}_{\taxis,r};\gb{\theta}_{\targ}+\gb{\theta}_{\targ,r}]))
\end{align}
where $\textbf{MLP}: \mathbb{R}^{2d}\rightarrow\mathbb{R}^{2d}$ is a multi-layer perceptron network, $[\cdot;\cdot]$ is the concatenation of two vectors, and $g$ is a function that generates $\gb{\theta}'_{\taxis}\in[-\pi, \pi)^d$ and $\gb{\theta}'_{\targ}\in[0,2\pi]^d$.
We define $g$ as:
\begin{align*}
[g(\textbf{x})]_i=
\begin{cases}
    \theta'^i_{\taxis}\ \ \ \ =\pi\tanh(\lambda_1 x_i), &\text{ if } i\le d,\\
    \theta'^{i-d}_{\targ}=\pi\tanh(\lambda_2 x_i)+\pi,&\text{ if } i> d.\\
\end{cases}
\end{align*}
where $[g(\textbf{x})]_i$ denotes the $i$-th element of $g(\textbf{x})$, $\lambda_1$ and $\lambda_2$ are two fixed parameters to control the scale.
Note that the range of the hyperbolic tangent function ($\tanh$) are open sets. Thus, we cannot indeed get the boundary value $\theta'^i_{\taxis}=-\pi$ and $\theta'^i_{\targ}=0,2\pi$. However, when we implement $g$ in experiments, the value of $g$ can be very close to 0 and 2$\pi$, which is equivalent to the closed set numerically.

\textbf{Intersection Operator} $\mathcal{I}$\textbf{.} Given a query $q$ that is the conjunction of conjunctive queries $q_i$, the goal of $\mathcal{I}$ is to represent $\llbracket q\rrbracket = \cap_{i=1}^n\llbracket q_i\rrbracket$. Since the conjunction of conjunctive queries are still conjunctive queries, the entities in $\llbracket q\rrbracket$ should have similar semantics. Recall that we only need to model entity sets that can be answers. We still use a Cartesian product of sector-cones to represent $\llbracket q\rrbracket$ (see Figure \ref{subfig:intersection}). 
Suppose that $\textbf{V}_q^c=(\gb{\theta}_{\taxis}, \gb{\theta}_{\targ})$ and $\textbf{V}_{q_i}^c=(\gb{\theta}_{i,\taxis}, \gb{\theta}_{i,\targ})$ are cone embeddings for $\llbracket q\rrbracket$ and $\llbracket q_i\rrbracket$, respectively. 
We define the intersection operator as follows:
\begin{align*}
    \gb{\theta}_{\taxis}&=\textbf{SemanticAverage}(\textbf{V}_{q_1}^c,\dots,\textbf{V}_{q_n}^c),\\
   \gb{\theta}_{\targ}&=\textbf{CardMin}(\textbf{V}_{q_1}^c,\dots,\textbf{V}_{q_n}^c),
\end{align*}
where $\textbf{SemanticAverage}(\cdot)$ and $\textbf{CardMin}(\cdot)$ generates semantic centers and apertures, respectively. In the following, we introduce these two functions in detail.

\textbf{\textit{SemanticAverage}.} As the semantic center of $\textbf{V}_q^c$, $\gb{\theta}_{\taxis}$ should be close to all the semantic centers $\gb{\theta}_{i,\taxis}$. Thus, we propose to represent $\gb{\theta}_{\taxis}$ as a semantic average of $\gb{\theta}_{i,\taxis}$. We note that the ordinary weighted average may lead to inconsistent semantics. For example, when $d=1$, if $\gb{\theta}_{1,\taxis}=\pi-\epsilon$ and $\gb{\theta}_{2,\taxis}=-\pi+\epsilon$ ($0<\epsilon<\pi/4$), then we expect $\gb{\theta}_{\taxis}$ to be around $\pi$. However, if we use the ordinary weighted sum, $\gb{\theta}_{\taxis}$ will be around $0$ with a high probability. To tackle this issue, we propose a semantic average scheme, which takes periodicity of axes into account. For a figure illustration of the difference between the ordinary and semantic average, please refer to Appendix D.

Specifically, we first map $[\gb{\theta}_{i,\taxis}]_j$ to points on the unit circle. Then, compute the weighted average of the points using an attention mechanism. Finally, map the points back to angles that represent axes. 
Formally, the computation process is 
\begin{align*}
    [\mathbf{x};\mathbf{y}]&=\sum_{i=1}^n [\textbf{a}_i\circ\cos(\gb{\theta}_{i,\taxis});\textbf{a}_i\circ\sin(\gb{\theta}_{i,\taxis})],\\
    \gb{\theta}_{\taxis}&=\textbf{Arg}(\mathbf{x},\mathbf{y}),
\end{align*}
where $\cos$ and $\sin$ are element-wise cosine and sine functions; $\textbf{a}_i\in\mathbb{R}^d$ are positive weights vectors that satisfy $\sum_{i=1}^n [\textbf{a}_i]_j=1$ for all $j=1,\dots,d$; $\circ$ is the element-wise multiplication; $\textbf{Arg}(\cdot)$ is the function that computes arguments of (a Cartesian of) 2D points. 
Noticing that the weights $\textbf{a}_i$ are relevant to both axes and apertures, we compute $[\textbf{a}_i]_j$ via the following attention mechanism:
\begin{align*}
    [\textbf{a}_i]_j=\frac{\exp([\textbf{MLP}([\gb{\theta}_{i,\taxis}-\gb{\theta}_{i,\targ}/2;\gb{\theta}_{i,\taxis}+\gb{\theta}_{i,\targ}/2])]_j)}{\sum_{k=1}^n\exp([\textbf{MLP}([\gb{\theta}_{k,\taxis}-\gb{\theta}_{k,\targ}/2;\gb{\theta}_{k,\taxis}+\gb{\theta}_{k,\targ}/2])]_j)},
\end{align*}
where $\textbf{MLP}:\mathbb{R}^{2d}\rightarrow\mathbb{R}^d$ is a multi-layer perceptron network, $[\cdot;\cdot]$ is the concatenation of two vectors.
We can see $\gb{\theta}_{i,\taxis}-\gb{\theta}_{i,\targ}/2$ and $\gb{\theta}_{i,\taxis}+\gb{\theta}_{i,\targ}/2$ as the lower and upper bound of sector-cones.

We use $\textbf{Arg}(\cdot)$ to recover angles of 2D points. Suppose that $\beta_i=\arctan([\textbf{y}]_i/[\textbf{x}]_i)$, then
\begin{align*}
[\gb{\theta}_{\taxis}]_i&=
    \begin{cases}
    \beta_i+\pi, &\text{ if }[\textbf{x}]_i<0, [\textbf{y}]_i>0, \\
    \beta_i-\pi, &\text{ if }[\textbf{x}]_i<0, [\textbf{y}]_i<0, \\
    \,\,\,\,\,\,\ \beta_i, &\text{otherwise}.
    \end{cases}
\end{align*}
Note that $[\textbf{x}]_i=0$ will lead to an illegal division. In experiments, we manually set $[\textbf{x}]_i$ to be a small number (e.g., $10^{-3}$) when $[\textbf{x}]_i=0$.

\textbf{\textit{CardMin}.} Since $\llbracket q\rrbracket$ is the subset of all $\llbracket q_i\rrbracket$, $\theta_{\targ}^i$ should be no larger than any apertures $\theta_{j,\targ}^i$. Therefore, we implement \textbf{CardMin} by a minimum mechanism with cardinality constraints: 
\begin{align*}
    \theta_{\targ}^i=\min\{{\theta}_{1,\targ}^i,\dots,{\theta}_{n,\targ}^i\}\cdot \sigma([\textbf{DeepSets}(\{\textbf{V}_{q_j}\}_{j=1}^n)]_i),
\end{align*}
where $\sigma(\cdot)$ is the element-wise sigmoid function, ${\theta}_{j,\targ}^i$ is the $i$-th element of $\gb{\theta}_{j,\targ}$, $\textbf{DeepSets}(\cdot)$ is a permutation-invariant function \cite{deepsets}. Specifically, $\textbf{DeepSets}(\{\textbf{V}_{q_j}\}_{j=1}^n)$ is computed by
\begin{align*}
    \textbf{MLP}\left(\frac{1}{n}\sum\nolimits_{j=1}^n\textbf{MLP}\left([\gb{\theta}_{j,\taxis}-\gb{\theta}_{j,\targ}/2;\gb{\theta}_{j,\taxis}+\gb{\theta}_{j,\targ}/2]\right)\right).
\end{align*}

\textbf{Union Operator} $\mathcal{U}$\textbf{.} Given a query $q$ that is the disjunction of conjunctive queries $q_i$, the goal of the union operator $\mathcal{U}$ is to represent $\llbracket q\rrbracket = \cup_{i=1}^n\llbracket q_i\rrbracket$. As noted by \citet{q2b}, directly modeling the disjunction leads to unscalable models. Thus, we adopt the DNF technique \cite{q2b}, in which the union operation only appears in the last step in computation graphs.

Suppose that $\textbf{V}_{q_i}^c=(\gb{\theta}_{i,\taxis}, \gb{\theta}_{i,\targ})$ are cone embeddings for $\llbracket q_i\rrbracket$. To represent the union of several cones (see Figure \ref{subfig:union}), we represent $\llbracket q\rrbracket$ as a set of $\textbf{V}_{q_i}^c$:
\begin{align*}
    \textbf{V}_{q}^d=\{\textbf{V}_{q_1}^c, \dots, \textbf{V}_{q_n}^c\},
\end{align*}
where $n$ may be various in different queries. Equivalently, $\textbf{V}_{q}^d$ can be written as 
\begin{align*}
    \textbf{V}_{q}^d=\left( \{({\theta}_{1,\taxis}^1, {\theta}_{1,\targ}^1), \dots, ({\theta}_{n,\taxis}^1, {\theta}_{n,\targ}^1)\}, \dots, \{({\theta}_{1,\taxis}^d, {\theta}_{1,\targ}^d), \dots, ({\theta}_{n,\taxis}^d, {\theta}_{n,\targ}^d)\} \right).
\end{align*}

As $\{({\theta}_{1,\taxis}^i, {\theta}_{1,\targ}^i), \dots, ({\theta}_{n,\taxis}^i, {\theta}_{n,\targ}^i)\}$ are the union of $d$ sector-cones, it is also a cone. Thus, the cone embedding of $q$ is also a Cartesian product of \textit{two-dimensional cones}.

\textbf{Complement Operator} $\mathcal{C}$\textbf{.} Given an conjunctive query $q$ and the corresponding entity set $\llbracket q\rrbracket$, the aim of $\mathcal{C}$ is to identify the set $\llbracket \neg q\rrbracket$, which is the complementary of $\llbracket q\rrbracket$, i.e., $\mathcal{V}\backslash \llbracket q\rrbracket$. Since the set of sector-cones is closed under closure-complement, we define $\mathcal{C}$ using the closure-complement. Thus, the apertures of $\textbf{V}_{q}$ plus the apertures of $\textbf{V}_{\neg q}$ should be a vector with all elements being $2\pi$. 
Moreover, to represent the semantic difference between $\llbracket q\rrbracket$ and $\llbracket \neg q\rrbracket$, we assume that their semantic centers to be opposite. Please refer to Figure \ref{subfig:complement} for a figure illustration.

Suppose that $\textbf{V}_q=(\gb{\theta}_{\taxis},\gb{\theta}_{\targ})$ and $\textbf{V}_{\neg q}=(\gb{\theta}'_{\taxis},\gb{\theta}'_{\targ})$. We define the complement operator $\mathcal{C}$ as:
\begin{align*}
   [\gb{\theta}'_{\taxis}]_i&=
    \begin{cases}
        [\gb{\theta}_{\taxis}]_i-\pi,&\text{ if }[\gb{\theta}_{\taxis}]_i\ge 0,\\
        [\gb{\theta}_{\taxis}]_i+\pi,&\text{ if }[\gb{\theta}_{\taxis}]_i< 0,
    \end{cases}\\
    [\gb{\theta}'_{\targ}]_i&=2\pi-[\gb{\theta}_{\targ}]_i.
\end{align*}
\subsection{Learning Cone Embeddings}\label{sec:train_obj}
\vspace{-1mm}
To learn cone embeddings, we expect that the cone embeddings of entities $v\in \llbracket q\rrbracket$ are inside the cone embeddings of $q$, and the cone embeddings of entities $v'\notin \llbracket q\rrbracket$ are far from the cone embedding of $q$. This motivates us to define a distance function to measure the distance between a given query embedding and an entity embedding, and a training objective with negative sampling.

\textbf{Distance Function.} We first define the distance function for conjunctive queries. Inspired by \citet{q2b}, we divide the distance $d$ into two parts---the outside distance $d_o$ and the inside distance $d_i$. Figure \ref{subfig:distance} gives an illustration of the distance function $d$. Suppose that $\textbf{v}=(\gb{\theta}^v_{\taxis},\textbf{0})$, $\textbf{V}_q^c=(\gb{\theta}_{\taxis},\gb{\theta}_{\targ})$, $\gb{\theta}_L=\gb{\theta}_{\taxis}-\gb{\theta}_{\targ}/2$ and $\gb{\theta}_U=\gb{\theta}_{\taxis}+\gb{\theta}_{\targ}/2$. We define the distance as
\begin{align*}
    d_{con}(\ee;\qe^c)=d_o(\ee;\qe^c)+\lambda d_i(\ee;\qe^c).
\end{align*}
The outside distance and the inside distance are 
\begin{align*}
    d_o&=\left\|\min\left\{\left|\sin\left(\gb{\theta}^v_{\taxis}-\gb{\theta}_L\right)/2\right|,\left|\sin\left(\gb{\theta}^v_{\taxis}-\gb{\theta}_U\right)/2\right|\right\}\right\|_1,\\
    d_i&=\left\|\min\left\{\left|\sin\left(\gb{\theta}^v_{\taxis}-\gb{\theta}_{\taxis}\right)/2\right|,\left|\sin\left(\gb{\theta}_{\targ}\right)/2\right| \right\}\right\|_1,
\end{align*}
where $\|\cdot\|_1$is the $L_1$ norm, $\sin(\cdot)$ and $\min(\cdot)$ are element-wise sine and minimization functions. Note that as axes and apertures are periodic, we use the sine function to enforce two close angles have small distance. The parameter $\lambda\in (0,1)$ is fixed during training, so that $\textbf{v}$ is encouraged to be inside the cones represented by $\textbf{V}_q^c$,
but not necessarily be equal to the semantic center of $\textbf{V}_q^c$.

Since we represent the disjunctive queries as a set of embeddings, we cannot use $d_{con}$ to directly compute the distance. Nonetheless, the distance between a point and the union of several sets is equal to the minimum distance between the point and each of those sets. Therefore, for a query $q=q_1\lor \dots\lor q_n$ in the Disjunctive Normal Form, the distance between $q$ and an entity is
\begin{align*}
    d_{dis}(\textbf{v};\textbf{V}_{q}^d)=\min\{d_{con}(\textbf{v};\textbf{V}_{q_1}^c),\dots,d_{con}(\textbf{v};\textbf{V}_{q_n}^c)\}.
\end{align*}
If we use $\textbf{V}_q$ to represent embeddings of both kinds of queries, the unified distance function $d$ is
\vspace{-1mm}
\begin{align*}
     d(\ee;\qe)=
     \begin{cases}
        d_{con}(\ee;\qe),\ \text{if }q \text{ is conjunctive queries},\\
        d_{dis}(\ee;\qe),\ \text{if }q \text{ is disjunctive queries}.
     \end{cases}
\end{align*}
\textbf{Training Objective.} Given a training set of queries, we optimize a negative sampling loss
\begin{align*}
    L=-\log\sigma(\gamma-d(\ee;\qe))-\frac{1}{k}\sum\nolimits_{i=1}^k\log\sigma(d(\ee'_i;\qe)-\gamma),
\end{align*}
where $\gamma>0$ is a fixed margin, $v\in\llbracket q\rrbracket$ is a positive entity, $v_i'\notin \llbracket q \rrbracket$ is the $i$-th negative entity, $k$ is the number of negative entities, and $\sigma(\cdot)$ is the sigmoid function.

\begin{table}[ht]
    \centering
    \caption{MRR results for answering queries without negation ($\exists$, $\land$, $\lor$) on FB15k, FB237, and NELL. The results of \betae~ are taken from \citet{beta}. }
    \vspace{1mm}
    \label{table:epfo_results}
    \setlength{\tabcolsep}{2mm}{
    \begin{tabular}{c c c c c c c c c c c c c c }
        \toprule
         \textbf{Dataset} & \textbf{Model} &\textbf{1p} &\textbf{2p} &\textbf{3p} &\textbf{2i} &\textbf{3i} &\textbf{pi} &\textbf{ip} &\textbf{2u} &\textbf{up} &\textbf{AVG}\\
        \midrule
        \multirow{4}{*}{FB15k}&GQE &53.9 &15.5 &11.1 &40.2 &52.4 &27.5 &19.4 &22.3 &11.7  &28.2\\
        &Q2B &70.5 &23.0 &15.1 &61.2 &71.8 &41.8 &28.7 &37.7 &19.0 &40.1\\
        &\betae~ &65.1 &25.7 &24.7 &55.8 &66.5 &43.9 &28.1 &40.1 &25.2 &41.6\\
        &ConE &\textbf{73.3} & \textbf{33.8} &\textbf{29.2} &\textbf{64.4} &\textbf{73.7} &\textbf{50.9} &\textbf{35.7} &\textbf{55.7} &\textbf{31.4}  &\textbf{49.8}\\
        \midrule
        \multirow{4}{*}{FB237}&GQE &35.2 &7.4 &5.5 &23.6 &35.7 &16.7 &10.9 &8.4 &5.8  &16.6\\
        &Q2B  &41.3 &9.9 &7.2 &31.1 &45.4 &21.9 &13.3 &11.9 &8.1  &21.1\\
        &\betae~ &39.0 &10.9 &10.0 &28.8 &42.5 &22.4 &12.6 &12.4 &9.7  &20.9\\
        &ConE &\textbf{41.8} &\textbf{12.8} &\textbf{11.0} &\textbf{32.6} &\textbf{47.3} &\textbf{25.5} &\textbf{14.0} &\textbf{14.5} &\textbf{10.8}  &\textbf{23.4}\\
        \midrule
        \multirow{4}{*}{NELL}&GQE &33.1 &12.1 &9.9 &27.3 &35.1 &18.5 &14.5 &8.5 &9.0 &18.7\\
        &Q2B &42.7 &14.5 &11.7 &34.7 &45.8 &23.2 &17.4 &12.0 &10.7  &23.6\\
        &\betae~ &53.0 &13.0 &11.4 &37.6 &47.5 &24.1 &14.3 &12.2 &8.5 &24.6\\
        &ConE &\textbf{53.1} &\textbf{16.1} &\textbf{13.9} &\textbf{40.0} &\textbf{50.8} &\textbf{26.3} &\textbf{17.5} &\textbf{15.3} &\textbf{11.3} &\textbf{27.2}\\
        \bottomrule
    \end{tabular}
    }
    \vspace{-4mm}
\end{table}

\vspace{-1mm}
\section{Experiments}
\vspace{-2mm}
In this section, we conduct experiments to demonstrate that: 1) ConE is a powerful model for the multi-hop reasoning over knowledge graphs; 2) the aperture embeddings of ConE are effective in modeling cardinality (i.e., the number of elements) of answer sets. We first introduce experimental settings in Section \ref{sec:exp_settings} and then present the experimental results in Sections \ref{sec:main_results} and \ref{sec:card_cor}. The code of ConE is available on GitHub at \url{https://github.com/MIRALab-USTC/QE-ConE}.
\vspace{-2mm}

\subsection{Experimental Settings}\label{sec:exp_settings}
\vspace{-1mm}
We adopt the commonly used experimental settings for query embeddings \cite{gqe,q2b,beta}. 

\textbf{Datasets and Queries.} We use three datasets: FB15k \cite{freebase}, FB15k-237 (FB237) \cite{fb237}, and NELL995 (NELL) \cite{deeppath}. 
QE models focus on answering queries involved with incomplete KGs. Thus, we aim to find non-trivial answers to FOL queries that cannot be discovered by traversing KGs. 
For a fair comparison, we use the same query structures as those in \citet{beta}. The training and validation queries consist of five conjunctive structures ($1p/2p/3p/2i/3i$) and five structures with negation ($2in/3in/inp/pni/pin$). We also evaluate models' generalization ability, i.e., answering queries with structures that models have never seen during training. The extra query structures include $ip/pi/2u/up$. Please refer to Appendix B.1 for more details about datasets and query structures.

\textbf{Training Protocol.} 
We use Adam \cite{adam} as the optimizer, and use grid search to find the best hyperparameters based on the performance on the validation datasets. For the search range and best hyperparameters, please refer to Appendix B.2.

\textbf{Evaluation Protocol.} We use the same evaluation protocol as that in \citet{beta}. We first build three KGs: the training KG $\mathcal{G}_{\ttrain}$, the validation KG $\mathcal{G}_{\tvalid}$, and the test KG $\mathcal{G}_{\ttest}$ using training edges, training+validation edges, training+validation+test edges, respectively. Given a test (validation) query $q$, we aim to discover non-trivial answers $\llbracket q\rrbracket_{\ttest}\backslash \llbracket q\rrbracket_{\tvalid}$ ($\llbracket q\rrbracket_{\tvalid}\backslash \llbracket q\rrbracket_{\ttrain})$. In other words, to answer an entity, we need to impute at least one edge to create an answer path to it. For each non-trivial answer $v$ of a test query $q$, we rank it against non-answer entities $\mathcal{V}\backslash \llbracket q\rrbracket_{\ttest}$. We denote the rank as $r$ and calculate the Mean Reciprocal Rank (MRR), of which the definition is provided in Appendix B.3. Higher MRR indicates better performance.

\textbf{Baselines.} We compare ConE against three state-of-the-art models, including GQE \cite{gqe}, Query2Box (Q2B) \cite{q2b}, and \betae~ \cite{beta}. GQE and Q2B are trained only on five conjunctive structures as they cannot model the queries with negation. Since the best embedding dimension $d$ for ConE is $800$, we retrain all the baselines with $d=800$. The results of GQE and Q2B are better than those reported in \citet{beta}, while the results of \betae~ become slightly worse. Therefore, we reported the results of GQE and Q2B with $d=1600$ and \betae~ with $d=400$. For the results of \betae~ with $d=800$, please refer to Appendix C.1.

\vspace{-1.5mm}
\subsection{Main Results}\label{sec:main_results}
\vspace{-1mm}
We compare ConE against baseline models on queries with and without negation. We run our model five times with different random seeds and report the average performance. For the error bars of the performance, please refer to Appendix C.5.

\textbf{Queries without Negation.}
Table \ref{table:epfo_results} shows the experimental results on queries without negation, i.e., existentially positive first-order (EPFO) queries, where \textbf{AVG} denotes average performance.
Overall, ConE significantly outperforms compared models. ConE achieves on average 19.7\%, 12.0\%, and 10.6\% relative improvement MRR over previous state-of-the-art \betae~ on the three datasets, which demonstrates the superiority of geometry-based models. 
Compared with Q2B, which uses Query2Box to embed queries, ConE gains up to 24.2\% relative improvements. ConE also gains an impressive improvement on queries $ip/pi/2u/up$, which are not in the training graph. For example, ConE outperforms \betae~ by 38.9\% for $2u$ query on FB15k. The results show the superior generality ability of ConE.
Since ConE is capable of modeling complement, we can also implement disjunctive queries using De Morgan's law. However, using De Morgan's law always results in sector-cones, which may be inconsistent with the real set union. Thus, the models with DNF outperforms those with De Morgan's law. We include the detailed results in Appendix C.2 due to the space limit.

\begin{table}[t]
    \centering
    \caption{MRR results for answering queries with negation on FB15k, FB237, and NELL. The results of \betae~ are taken from \citet{beta}. }
    \label{table:negation_results}
    \vspace{1mm}
    \begin{tabular}{ c c  c c c c  c c c c }
        \toprule
         \textbf{Dataset} & \textbf{Model} &\textbf{2in} &\textbf{3in} &\textbf{inp} &\textbf{pin} &\textbf{pni} &\textbf{AVG}\\
        \midrule
        \multirow{2}{*}{FB15k} &\betae~ &14.3 &14.7 &11.5 &6.5 &12.4 &11.8\\
        &ConE &\textbf{17.9} &\textbf{18.7} &\textbf{12.5} &\textbf{9.8} &\textbf{15.1} &\textbf{14.8}\\
        \midrule
        \multirow{2}{*}{FB237}&\betae~ &5.1 &7.9 &7.4 &3.6 &3.4 &5.4\\
        &ConE &\textbf{5.4} &\textbf{8.6} &\textbf{7.8} &\textbf{4.0} &\textbf{3.6} &\textbf{5.9}\\
        \midrule
        \multirow{2}{*}{NELL} &\betae~ &5.1 &7.8 &10.0 &3.1 &3.5 &5.9\\
        &ConE &\textbf{5.7} &\textbf{8.1} &\textbf{10.8} &\textbf{3.5} &\textbf{3.9} &\textbf{6.4}\\
        \bottomrule
    \end{tabular}
    \vspace{-3mm}
\end{table}

\textbf{Queries with Negation.}
Table \ref{table:negation_results} shows the results of ConE against \betae~ on modeling FOL queries with negation. Since GQE and Q2B are not capable of handling the negation operator, we do not include their results in the experiments. Overall, ConE outperforms \betae~ by a large margin. Specifically, ConE achieves on average 25.4\%, 9.3\%, and 8.5\% relative improvement MRR over \betae~ on FB15k, FB237, and NELL, respectively. 

\begin{table}[ht]
    \centering
    \caption{Spearman's rank correlation between learned aperture embeddings and the number of queries' answers on FB15k. The results of Q2B and \betae~ are taken from \citet{beta}. }
    \label{table:spearman}
    \vspace{1mm}
    \setlength{\tabcolsep}{2.0mm}{
    \begin{tabular}{c c c  c c c c  c c c c c c c}
        \toprule
         \textbf{Model} &\textbf{1p} &\textbf{2p} &\textbf{3p} &\textbf{2i} &\textbf{3i} &\textbf{pi} &\textbf{ip} &\textbf{2in} &\textbf{3in} &\textbf{inp} &\textbf{pin} &\textbf{pni} \\
        \midrule
        Q2B &0.30 &0.22 &0.26 &0.33 &0.27 &0.30 &0.14 &- &- &- &- &-\\
        \betae~ &0.37 &0.48 &0.47 &0.57 &0.40 &0.52 &0.42 &0.62 &0.55 &0.46 & 0.47 &0.61\\
        ConE &\textbf{0.60} & \textbf{0.68} &\textbf{0.70} &\textbf{0.68} &\textbf{0.52} &\textbf{0.59} &\textbf{0.56} &\textbf{0.84} &\textbf{0.75} &\textbf{0.61} &\textbf{0.58} &\textbf{0.80} \\
        \bottomrule
    \end{tabular}
    }
\end{table}
\vspace{-1mm}
\subsection{Modeling the Cardinality of Answer Sets}\label{sec:card_cor}
\vspace{-1mm}
As introduced in Section \ref{sec:cone_e}, the aperture embeddings can designate the cardinality (i.e., the number of elements) of $\llbracket q\rrbracket$. In this experiment, we demonstrate that although we do not explicitly enforce ConE to learn cardinality during training, the learned aperture embeddings are effective in modeling the cardinality of answer sets. The property partly accounts for the empirical improvements of ConE.

We compute the correlations between learned aperture embeddings and the cardinality of answer sets. Specifically, for the cone embedding $\textbf{V}_q=(\gb{\theta}_{\taxis}, \gb{\theta}_{\targ})$ of a given query $q$, we use the $L_1$ norm of $\gb{\theta}_{\targ}$ to represent the learned cardinality of $\llbracket q\rrbracket$. Then, we compute the Spearman's rank correlation (SRC) between the learned cardinality and the real cardinality, which measures the statistical dependence between the ranking of two variables. Higher correlation indicates that the embeddings can better model the cardinality of answer sets. As we model queries with disjunction using the DNF technique, we do not include the results of disjunctive queries following \citet{beta}.

Table \ref{table:spearman} shows the results of SRC for ConE, Query2Box (Q2B), and \betae~ on FB15k. For the results on FB237 and NELL, please refer to Appendix C.3. As Query2Box cannot handle queries with negation, we do not include its results on these queries. On all query structures, ConE outperforms the previous state-of-the-art method \betae. Note that \betae~ is a probabilistic model, of which the authors claim that it can well handle the uncertainty of queries, i.e., the cardinality of answer set. Nonetheless, ConE still outperforms \betae~ by a large margin, which demonstrates the expressiveness of cone embeddings.
We also conduct experiments using Pearson's correlation, which measures the linear correlation between two variables. Please refer to Appendix C.3 for the results.

\subsection{The Designed Operators and the Real Set Operations }
As introduced in Section \ref{sec:logical_op}, the designed union (using DNF technique) and complement operators for ConE are non-parametric. They correspond to \textit{exact} set union and complement. Meanwhile, we define neural operators to approximate the projection and intersection operators to achieve a tractable training process and better performance. Notably, the designed neural operators may not exactly match the real set operations. However, experiments on some example cases demonstrate that these neural operators provide good approximations for real set operations. In the following, we show the experimental results for the operators including projection and intersection. In all experiments, the ConE embeddings are trained on FB15k. 

\udfsection{Projection.} 
Suppose that a set $A$ is included by a set $B$, then we expect the projection of $A$ is also included by the projection of $B$. We randomly generate 8000 pairs of sector-cones $(A_i, B_i)$, where $A_i\subset B_i$. Then, for each $i$, we randomly select a relation $r_i$ and calculate the projections $P_{r_i}(A_i)$ and $P_{r_i}(B_i)$. Ideally, the projected cones should satisfy $P_{r_i}(A_i)\subset P_{r_i}(B_i)$. We calculate the ratio $r_i=|P_{r_i}(A_i)\cap P_{r_i}(B_i)|/|P_{r_i}(A_i)|$ to measure how many elements in $P_{r_i}(A_i)$ are included in $P_{r_i}(A_i)\cap P_{r_i}(B_i)$. Finally, we get an average ratio $r=0.8113$. That is to say, the learned  $P_{r_i}(A_i)$ are included in $P_{r_i}(A_i)$ with a high probability. The learned projection operators approximate the real set projection well.

\udfsection{Intersection.}
To validate that the learned intersection can well approximate real set intersection, we randomly generate 8000 pairs of sector-cones $(C_i, D_i)$, where $C_i\cap D_i$ is not guaranteed to be a sector-cone. Then, we generate embeddings $I(C_i, D_i)$ for the intersection $C_i\cap D_i$. Ideally, the learned cones $I(C_i, D_i)$ should be the same as the real cones $C_i\cap D_i$. We calculate the ratio $r_i=|I(C_i, D_i)\cap (C_i\cap D_i)|/|I(C_i, D_i)\cup (C_i\cap D_i)|$ to measure the overlap between $I(C_i, D_i)$ and $C_i\cap D_i$, and obtain an average ratio of $r=0.6134$. Note that the experiments are conducted on the test set, and we did not explicitly train our model on these queries. The relatively high overlap ratio $r$ demonstrates that the learned intersection is a good approximation of the real set intersection.

We further conduct experiments to demonstrate that the learned intersection operators can well handle empty intersections. Following \citet{q2b}, on FB15k, we randomly generate $10k$ queries of two types: (a) intersection queries with more than five answers, and (b) intersection queries with empty answer sets. We found that the average aperture is $2.495$ for type (a) queries, while $1.737$ for type (b) queries. The results demonstrate that although we have never trained ConE on the type (b) queries, the empty intersection sets are much more likely to have smaller apertures than queries with non-zero answers (with a $0.9632$ ROC-AUC score). In other words, though we did not train ConE on datasets with empty intersection sets, we can distinguish empty answer sets by the learned apertures.

We also conduct experiments to demonstrate the difference between the learned union operator with De Morgan's law and the real set union. Please refer to Appendix C.6 for details.

\vspace{-1mm}
\section{Conclusion}\label{sec:conclusion}
\vspace{-2mm}
In this paper, we propose a novel query embedding model, namely Cone Embeddings (ConE), to answer multi-hop first-order logical (FOL) queries over knowledge graphs.
We represent entity sets as Cartesian products of cones and design corresponding logical operations. To the best of our knowledge, ConE is the first geometric query embedding models that can model all the FOL operations. Experiments demonstrate that ConE significantly outperforms previous state-of-the-art models on benchmark datasets. One future direction is to adapt ConE to queries in the natural language, which will further improve ConE's applicability.

\bibliography{neurips_2021}
\bibliographystyle{neurips_2021}

\newpage
\appendix
\section{Proof for Proposition 1}
To show Proposition 1, we need the following defition and lemma.

\begin{definition}[\citet{boyd}]
A cone $C\subset \mathbb{R}^2$ is called a proper cone if it satisfies: 
\begin{itemize}[leftmargin=10mm]
    \item $C$ is convex,
    \item $C$ is closed,
    \item $C$ is solid, which means it has nonempty interior,
    \item $C$ is pointed, which means that it contains no line (or equivalently, $x\in C, -x\in C\Rightarrow x=0$).
\end{itemize}
\end{definition}
\vspace{3mm}

\begin{lemma}[\citet{lemma_sym}]\label{lemma:sym_cone}
Suppose that $K$ is a proper cone. Then
\begin{align*}
    (n-1)^{-1}\le \text{as}(K)\le 1,
\end{align*}
where $\text{as}(K)$ is the axial symmetry degree of $K$. 
The upper bound becomes an equality if and only if $K$ is axially symmetric.
\end{lemma}
Be letting $n=2$ in Lemma \ref{lemma:sym_cone}, we know that $\text{as}(K)=1$, which attains the upper bound. Thus, proper cones in $\mathbb{R}^2$ are always axially symmetric.

\begin{prop}\label{appendix_prop:axial}
A sector-cone is always axially  symmetric.
\end{prop}
\vspace{-5mm}
\begin{proof}
Suppose that $C\subset\mathbb{R}^2$ is a sector-cone, then it is a closed cone. 

We further assume that $C$ is convex, contains no line, and has nonempty interior, i.e., it is a proper cone. By Lemma \ref{lemma:sym_cone}, we know that $C$ is axially symmetric. If $C$ is convex but contains a line, i.e., it is the half space, then it is axially symmetric. If $C$ has empty interior, i.e., it is a ray, then it is axially symmetric. Therefore, when $C$ is convex, it is axially symmetric.

If $C$ is not convex, then by the definition of sector-cones, its complement is convex. We know that the closure-complement of $C$, i.e., $\tilde{C}$, is a closed convex cone, and thus axially symmetric. It is easy to see that the axis of symmetry of $\tilde{C}$ is also the axis of symmetry of $C$, and $C$ is axially symmetric.

Therefore, a sector-cone is always axially symmetric.
\end{proof}

\begin{table}[ht]
    \caption{Statistics of three benchmark datasets, where FB237 denotes FB15k-237, EPFO represents $1p/2p/3p/2i/3i$, and $n1p$ represents $2in/3in/inp/pin/pni$.}
    \vspace{1mm}
    \label{table:datasets}
    \centering
    
    \begin{tabular}{c *{6}{c}}
        \toprule
         & \multicolumn{2}{c}{\textbf{Training}} & \multicolumn{2}{c}{\textbf{Validation}} & \multicolumn{2}{c}{\textbf{Test}} \\
         \cmidrule(lr){2-3}\cmidrule(lr){4-5}\cmidrule(lr){6-7}
        \textbf{Dataset} & \textbf{EPFO} &\textbf{Neg} & \textbf{1p} &\textbf{n1p} & \textbf{1p} &\textbf{n1p}\\
        \midrule
        FB15k & 273,710 &27,371 & 59,078 & 8,000 &66,990 & 8,000\\
        FB237 & 149,689 &14,968 & 20,094 & 5,000 &22,804 & 5,000\\
        NELL & 107,982 &10,798 & 16,910 & 4,000 &17,021 & 4,000\\
        \bottomrule
    \end{tabular}
\end{table}

\begin{figure*}[ht]
    \centering
    \includegraphics[width=0.9\columnwidth]{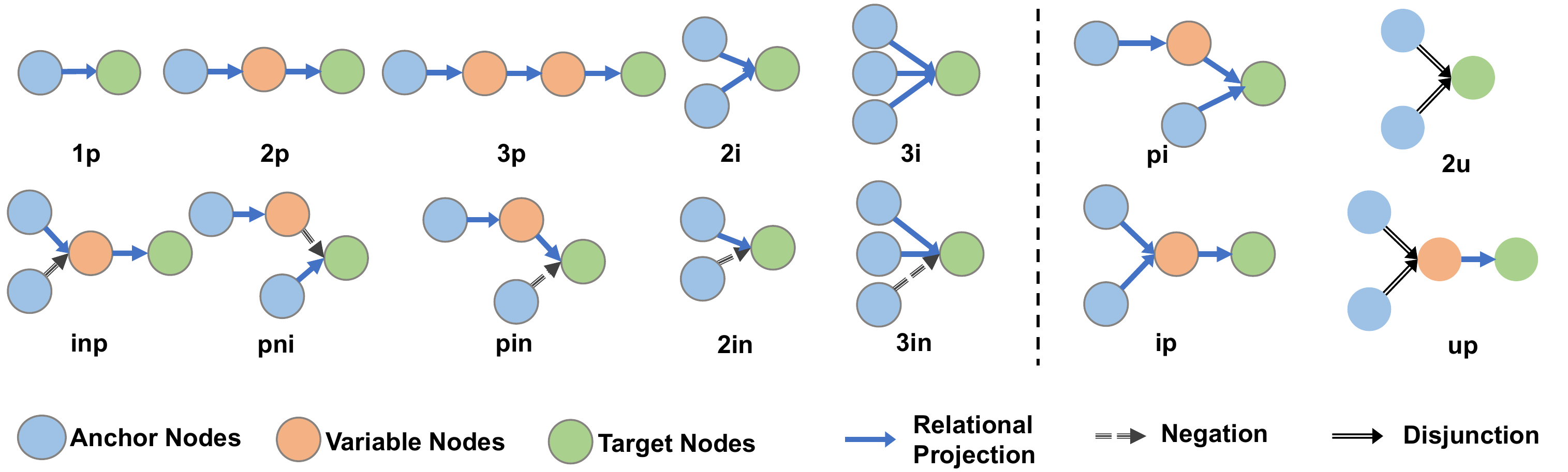}
    \caption{Fourteen queries used in the experiments. where $``p"$ denotes relation projection, $``i"$ denotes intersection, $``u"$ denotes union, and $``n"$ denotes negation. The left part of queries is used in the training phase. Both parts are used in the validation and test phases.}
    \vspace{1mm}
    \label{fig:queries}
    \vspace{-3mm}
\end{figure*}

\begin{table*}[ht]
    \centering
    \caption{Hyperparameters found by grid search. $d$ is the embedding dimension, $b$ is the batch size, $n$ is the negative sampling size, $\gamma$ is the parameter in the loss function, $m$ is the maximum training step, $l$ is the learning rate, $dr$ is the dropout rate for CardMin, $\lambda_1$ and $\lambda_2$ are weights in the projection operator, $\lambda$ is the parameter in the distance function $d_{con}$.}
    \vspace{1mm}
    \label{table:hp}
    \begin{tabular}{l  c c c c c  c c c c c }
    \toprule
        &$d$ &$b$ &$n$ &$\gamma$ &$m$ &$l$ &$dr$ &$\lambda_1$ &$\lambda_2$ &$\lambda$\\
        \midrule
        FB15k &800 &128 &512 &$300k$ &30 &$5\times 10^{-5}$ &0.05   &1.0 &2.0 &0.02\\
        FB237 &800 &128 &512 &$300k$ &30 &$1\times 10^{-4}$ &0.10   &1.0 &2.0 &0.02\\
        NELL  &800 &128 &512 &$450k$ &20 &$1\times 10^{-4}$ &0.20   &1.0 &2.0 &0.02\\
        \bottomrule
    \end{tabular}
\end{table*}

\section{More Details about Experiments}
In this section, we show more details about experiments that are not included in the main text due to the limited space.

\subsection{Datasets and Query Structures}
For a fair comparison, we use the same datasets and query structures as those in \citet{beta}. The datasets is created by \citet{beta} based on two well-known knowledge graphs Freebase \cite{freebase} and NELL \cite{nell}. They do not contain personally identifiable information or offensive content.  Table \ref{table:datasets} summarizes the number of different queries in different datasets. Figure \ref{fig:queries} shows all the query structures used in the experiments.

\subsection{Training Protocal}
We run all the experiments on a single Nvidia Geforce RTX 3090 GPU card. All the models are implemented in Pytorch \cite{pytorch} and based on the official implementation of \betae~\cite{beta}\footnote{Link: \url{https://github.com/snap-stanford/KGReasoning}, licensed under the MIT License.} for a fair comparison. We search the learning rates in $\{5\times 10^{-5}, 10^{-4}, 5\times 10^{-4}\}$, the batch size in $\{128,256,512\}$, the embedding size in $\{200,400,800\}$, the negative sample sizes in $\{32,64,128\}$, and the margin $\gamma$ in $\{20,30,40,50,60\}$. For all the modules using multi-layer perceptron (MLP), we use a three-layer MLP with 1600 hidden neurons and ReLU activation. We apply dropout to the $\min$ function in $\textbf{CardMin}$ and search the dropout rate in $\{0.05,0.10,0.15,0.20\}$. The best hyperparameters are shown in Table \ref{table:hp}.

\subsection{Evaluation Metrics}
We choose Mean Reciprocal Rank (MRR) as the evaluation metric. Higher MRR  indicates better performance. Definitions are as follows. The mean reciprocal rank is the average of the reciprocal ranks of results for a sample of queries Q:
\begin{align*}
    \text{MRR}=\frac{1}{|Q|}\sum_{i=1}^{|Q|}\frac{1}{\text{rank}_i}.
\end{align*}

\section{More Experimental Results}
In this section, we give more experimental results that are not included in the main text due to the limited space.

\subsection{Results of \betae~ with Embedding Dimension 800}
Tables \ref{table:epfo_results_800} and \ref{table:negation_results_800} show the results of \betae~ with embedding dimensions 400 (B-400) and 800 (B-800). The results of B-400 is slightly better than that of B-800. Therefore, we report the results of B-400 in the main text.

\vspace{5mm}
\begin{table*}[ht]
    \centering
    \caption{MRR results for answering queries without negation ($\exists$, $\land$, $\lor$) on FB15k, FB237, and NELL, where B-400 and B-800 denote \betae~ with embedding dimensions 400 and 800, respectively.
    The results of B-400 models are taken from \citet{beta}.}
    \label{table:epfo_results_800}
    \vspace{1mm}
    \begin{tabular}{c c c c c c c c c c c c }
        \toprule
         \textbf{Dataset} & \textbf{Model} &\textbf{1p} &\textbf{2p} &\textbf{3p} &\textbf{2i} &\textbf{3i} &\textbf{pi} &\textbf{ip} &\textbf{2u} &\textbf{up} &\textbf{AVG}\\
        \midrule
        \multirow{3}{*}{FB15k}
        &B-400 &65.1 &25.7 &24.7 &55.8 &66.5 &43.9 &28.1 &40.1 &25.2&41.6\\
        &B-800 &61.9 &25.1 &24.2 &56.5 &67.9 &43.7 &26.6 &38.8 &24.5  &41.0\\
        &ConE &\textbf{73.3} & \textbf{33.8} &\textbf{29.2} &\textbf{64.4} &\textbf{73.7} &\textbf{50.9} &\textbf{35.7} &\textbf{55.7} &\textbf{31.4}  &\textbf{49.8}\\
        \midrule
        \multirow{3}{*}{FB237}
        &B-400 &39.0 &10.9 &10.0 &28.8 &42.5 &22.4 &12.6 &12.4 &9.7 &20.9\\
        &B-800 &38.3 &10.6 &9.9 &28.5 &42.7 &21.9 &11.8 &11.9 &9.5  &20.6\\
        &ConE &\textbf{41.8} &\textbf{12.8} &\textbf{11.0} &\textbf{32.6} &\textbf{47.3} &\textbf{25.5} &\textbf{14.0} &\textbf{14.5} &\textbf{10.8}  &\textbf{23.4}\\
        \midrule
        \multirow{3}{*}{NELL}
        &B-400 &53.0 &13.0 &11.4 &37.6 &47.5 &24.1 &14.3 &12.2 &8.5 &24.6\\
        &B-800 &51.6 &12.5 &10.7 &36.9 &48.2 &23.3 &13.9 &11.8 &8.1  &24.1\\
        &ConE &\textbf{53.1} &\textbf{16.1} &\textbf{13.9} &\textbf{40.0} &\textbf{50.8} &\textbf{26.3} &\textbf{17.5} &\textbf{15.3} &\textbf{11.3} &\textbf{27.2}\\
        \bottomrule
    \end{tabular}
    \vspace{-1mm}
\end{table*}

\vspace{5mm}
\begin{table}[ht]
    \centering
    \caption{MRR results for answering queries with negation on FB15k, FB237, and NELL, where B-400 and B-800 denote \betae~ with embedding dimensions 400 and 800, respectively. The results of B-400 models are taken from \citet{beta}.}
    \label{table:negation_results_800}
    \vspace{1mm}
    \begin{tabular}{ c c  c c c c  c c c c }
        \toprule
         \textbf{Dataset} & \textbf{Model} &\textbf{2in} &\textbf{3in} &\textbf{inp} &\textbf{pin} &\textbf{pni} &\textbf{AVG}\\
        \midrule
        \multirow{3}{*}{FB15k} 
        &B-400 &14.3 &14.7 &11.5 &6.5 &12.4 &11.8\\
        &B-800 &13.7 &14.6 &11.3 &6.4 &11.9 &11.6\\
        &ConE &\textbf{18.6} &\textbf{19.4} &\textbf{12.6} &\textbf{10.0} &\textbf{15.4} &\textbf{15.2}\\
        \midrule
        \multirow{3}{*}{FB237}
        &B-400 &5.1 &7.9 &7.4 &3.6 &3.4 &5.4\\
        &B-800 &4.9 &7.1 &7.6 &3.7 &3.3 &5.3\\
        &ConE &\textbf{5.8} &\textbf{8.8} &\textbf{7.6} &\textbf{4.3} &\textbf{4.1} &\textbf{6.1}\\
        \midrule
        \multirow{3}{*}{NELL} 
        &B-400 &5.1 &7.8 &10.0 &3.1 &3.5 &5.9\\
        &B-800 &5.0 &7.7 &10.1 &3.1 &2.9 &5.7\\
        &ConE &\textbf{5.6} &\textbf{8.1} &\textbf{10.9} &\textbf{3.5} &\textbf{3.9} &\textbf{6.4}\\
        \bottomrule
    \end{tabular}
    \vspace{-5mm}
\end{table}

\newpage
\subsection{Results on Disjunctive Queries}
Since ConE is capable of modeling complement, we can also implement disjunctive queries using De Morgan's law, i.e., $\cup_{i=1}^n S_i=\overline{\cap_{i=1}^n\overline{S}_i}$. Table \ref{table:dis} shows the results of $2u$ and $up$ queries that are implemented using both DNF ($\textbf{-N}$) and De Morgan's law ($\textbf{-M}$). The results show that  results of $\text{2u-M}/\text{up-M}$ are competitive compared with those of $\text{2u-N}/\text{up-N}$, which all outperform \betae~.

We can also see that ConE using De Morgan's law perform worse than ConE using DNF. The results is reasonable and expectable.
If we use the complement to handle queries with unions, their representations will always be sector-cones. However, not all such queries can be well represented by sector-cones (see Figure 3c in the main text).

\begin{table}[ht]
    \centering
    \caption{MRR results for answering disjunctive queries on FB15k, FB237, and NELL. The results of \betae~ are taken from \citet{beta}. $\textbf{2u-N}/\textbf{up-N}$ indicates that the disjunction is implemented using the DNF technique, while $\textbf{2u-M}/\textbf{up-M}$ indicates the implementation using De Morgan's law.}
    \label{table:dis}
    \vspace{1mm}
    \begin{tabular}{ c c c  c c c }
        \toprule
         \textbf{Dataset}  & \textbf{Model} & \textbf{2u-N} &\textbf{2u-M} &\textbf{up-N} &\textbf{up-M}\\
        \midrule
        \multirow{2}{*}{FB15k} 
        &\betae &40.1 &25.0 &25.2 &25.4 \\
        &ConE &\textbf{55.7} &\textbf{37.7} &\textbf{31.4} &\textbf{29.8}\\
        \midrule
        \multirow{2}{*}{FB237}
        &\betae &12.4 &11.1 &9.7 &\textbf{9.9} \\
        &ConE &\textbf{14.5} &\textbf{13.4} &\textbf{10.8} &\textbf{9.9}\\
        \midrule
        \multirow{2}{*}{NELL} 
        &\betae &12.2 &11.0 &8.5 &8.6 \\
        &ConE &\textbf{15.3} &\textbf{14.8} &\textbf{11.3} &\textbf{10.8}\\
        \bottomrule
    \end{tabular}
    \vspace{-3mm}
\end{table}

\subsection{Correlation Results}
Tables \ref{table:spearman_fb237} and \ref{table:spearman_nell} show the results of Spearman's rank correlation between learned embeddings and the number of queries on FB15k-237 and NELL, respectively.  The results of Query2Box (Q2B) and \betae~ are taken from \citet{beta}. The symbol ``$*$" indicates that the average performance is computed only using results of queries without negation.

Tables  \ref{table:pearson_fb}, \ref{table:pearson_fb237} and \ref{table:pearson_nell} show the results of Pearson correlation between learned embeddings and the number of queries on FB15k, FB15k-237, and NELL, respectively.  The results of Query2Box (Q2B) and \betae~ are taken from \citet{beta}. The symbol ``$*$" indicates that the average performance is computed only using results of queries without negation.

All the results show that ConE is effective in modeling the cardinality of queries' answer sets.

\begin{table*}[ht]
    \centering
    \caption{Spearman's rank correlation between learned embeddings and the number of answers  on FB15k-237.}
    \label{table:spearman_fb237}
    \vspace{1mm}
    \setlength{\tabcolsep}{2.0mm}{
    \begin{tabular}{c c c  c c c c  c c c c c c c}
        \toprule
         \textbf{Model} &\textbf{1p} &\textbf{2p} &\textbf{3p} &\textbf{2i} &\textbf{3i} &\textbf{pi} &\textbf{ip} &\textbf{2in} &\textbf{3in} &\textbf{inp} &\textbf{pin} &\textbf{pni} \\
        \midrule
        Q2B &0.18 &0.23 &0.27 &0.35 &0.44 &0.36 &0.20 &- &- &- &- &- \\
        \betae~ &0.406 &0.50 &0.57 &0.60 &0.52 &0.54 &0.44 &0.69 &0.58 &0.51 & 0.47 &0.67 \\
        ConE &\textbf{0.70} & \textbf{0.71} &\textbf{0.74} &\textbf{0.82} &\textbf{0.72} &\textbf{0.70} &\textbf{0.62} &\textbf{0.90} &\textbf{0.83} &\textbf{0.66} &\textbf{0.57} &\textbf{0.88} \\
        \bottomrule
    \end{tabular}
    }
    \vspace{-1mm}
\end{table*}

\begin{table*}[!ht]
    \centering
    \caption{Spearman's rank correlation between learned embeddings and the number of answers on NELL.}
    \label{table:spearman_nell}
    \vspace{1mm}
    \begin{tabular}{c c c  c c c c  c c c c c c c}
        \toprule
         \textbf{Model} &\textbf{1p} &\textbf{2p} &\textbf{3p} &\textbf{2i} &\textbf{3i} &\textbf{pi} &\textbf{ip} &\textbf{2in} &\textbf{3in} &\textbf{inp} &\textbf{pin} &\textbf{pni}\\
        \midrule
        Q2B &0.15 &0.29 &0.31 &0.38 &0.41 &0.36 &0.35 &- &- &- &- &- \\
        \betae~ &0.42 &0.55 &0.56 &0.59 &0.61 &0.60 &0.54 &0.71 &0.60 &0.35 & 0.45 &0.64\\
        ConE &\textbf{0.56} & \textbf{0.61} &\textbf{0.60} &\textbf{0.79} &\textbf{0.79} &\textbf{0.74} &\textbf{0.58} &\textbf{0.90} &\textbf{0.79} &\textbf{0.56} &\textbf{0.48} &\textbf{0.85} \\
        \bottomrule
    \end{tabular}
    \vspace{-1mm}
\end{table*}

\begin{table*}[!ht]
    \centering
    \caption{Pearson correlation between learned embeddings and the number of answers on FB15k.}
    \label{table:pearson_fb}
    \vspace{1mm}
    \begin{tabular}{c c c  c c c c  c c c c c c c}
        \toprule
         \textbf{Model} &\textbf{1p} &\textbf{2p} &\textbf{3p} &\textbf{2i} &\textbf{3i} &\textbf{pi} &\textbf{ip} &\textbf{2in} &\textbf{3in} &\textbf{inp} &\textbf{pin} &\textbf{pni} \\
        \midrule
        Q2B &0.08 &0.22 &0.26 &0.29 &0.23 &0.25 &0.13 &- &- &- &- &- \\
        \betae~ &0.22 &0.36 &0.38 &0.39 &0.30 &0.31 &0.31 &0.44 &0.41 &0.34 & 0.36 &0.44\\
        ConE &\textbf{0.33} &\textbf{0.53} &\textbf{0.59} &\textbf{0.5} &\textbf{0.45} &\textbf{0.37} &\textbf{0.42} &\textbf{0.65} &\textbf{0.55} &\textbf{0.50} & \textbf{0.52} &\textbf{0.64} \\
        \bottomrule
    \end{tabular}
   \vspace{-1mm}
\end{table*}

\begin{table*}[!ht]
    \centering
    \caption{Pearson correlation between learned embeddings and the number of answers on FB15k-237.}
    \label{table:pearson_fb237}
    \vspace{1mm}
    \begin{tabular}{c c c  c c c c  c c c c c c c c}
        \toprule
         \textbf{Model} &\textbf{1p} &\textbf{2p} &\textbf{3p} &\textbf{2i} &\textbf{3i} &\textbf{pi} &\textbf{ip} &\textbf{2in} &\textbf{3in} &\textbf{inp} &\textbf{pin} &\textbf{pni} \\
        \midrule
        Q2B &0.02 &0.19 &0.26 &0.37 &0.49 &0.34 &0.20 &- &- &- &- &-\\
        \betae~ &0.23 &0.37 &0.45 &0.36 &0.31 &0.32 &0.33 &0.46 &0.41 &0.39 & 0.36 &0.48 \\
        ConE &\textbf{0.40} & \textbf{0.52} &\textbf{0.61} &\textbf{0.67} &\textbf{0.69} &\textbf{0.47} &\textbf{0.49} &\textbf{0.71} &\textbf{0.66} &\textbf{0.53} &\textbf{0.47} &\textbf{0.72} \\
        \bottomrule
    \end{tabular}
\end{table*}

\begin{table*}[!ht]
    \centering
    \caption{Pearson correlation between learned embeddings and the number of answers on NELL. }
    \label{table:pearson_nell}
    \vspace{1mm}
    \begin{tabular}{c c c  c c c c  c c c c c c c c}
        \toprule
         \textbf{Model} &\textbf{1p} &\textbf{2p} &\textbf{3p} &\textbf{2i} &\textbf{3i} &\textbf{pi} &\textbf{ip} &\textbf{2in} &\textbf{3in} &\textbf{inp} &\textbf{pin} &\textbf{pni} \\
        \midrule
        Q2B &0.07 &0.21 &0.31 &0.36 &0.29 &0.24 &0.34 &- &- &- &- &- \\
        \betae~ &0.24 &0.40 &0.43 &0.40 &0.39 &0.40 &\textbf{0.40} &0.52 &0.51 &0.26 & \textbf{0.35} &0.46  \\
        ConE &\textbf{0.48} & \textbf{0.45} &\textbf{0.49} &\textbf{0.72} &\textbf{0.68} &\textbf{0.52} &0.39 &\textbf{0.74} &\textbf{0.66} &\textbf{0.38} & 0.34 &\textbf{0.69} \\
        \bottomrule
    \end{tabular}
    \vspace{1mm}
\end{table*}

\subsection{Comparison with EmQL}
We compare ConE with EmQL \cite{faith} on FB15k that is from Query2Box \cite{q2b}. The dataset is the same as that in EmQL.
Table \ref{table:cmp_results} shows that ConE significantly outperforms EmQL and other baselines. 

\begin{table}[!h]
    \centering
    \caption{Comparison with EmQL \cite{faith} under the ``generalization" setting. The used dataset FB15k is from Query2Box \cite{q2b}, which is the same as that in EmQL. }
    \vspace{1mm}
    \label{table:cmp_results}
        \begin{tabular}{l c c c  c c }
            \toprule
          &GQE & Q2B & BetaE & EmQL & ConE \\
            \midrule
           AVG MRR & 33.2 & 41.0 & 44.6 & 43.9 & \textbf{52.9}\\
            \bottomrule
        \end{tabular}
\end{table}

\subsection{Error Bars of Main Results}
To evaluate the multi-hop reasoning performance of ConE, we run the model five times with random seeds $\{0,10,100,1000,10000\}$. In this section, we report the error bars of these results. Table \ref{table:errorbar_epfo} shows the error bar of ConE's MRR results on EPFO queries, i.e., queries without negation. Table \ref{table:errorbar_neg} shows the error bar of ConE's MRR results on queries with negation. Overall, the standard variances are small, which demonstrate that the performence of ConE is stable.

\begin{table*}[!ht]
    \centering
    \caption{The mean values and standard variances of ConE's MRR results on EPFO queries.}
    \label{table:errorbar_epfo}
    \vspace{1mm}
    \begin{tabular}{c c c  c c c c  c c c c c}
        \toprule
          \textbf{Dataset} &\textbf{1p} &\textbf{2p} &\textbf{3p} &\textbf{2i} &\textbf{3i} &\textbf{pi} &\textbf{ip} &\textbf{2u} &\textbf{up} &\textbf{AVG} \\
        \midrule
        \multirow{2}{*}{FB} &73.3 &33.8 &29.2 &64.4 &73.7 &50.9 &35.7 &55.7 &31.4 &49.8\\
        &\std{0.086} &\std{0.193} &\std{0.198} &\std{0.176} &\std{0.207} &\std{0.155} &\std{0.126} &\std{0.445} &\std{0.251} &\std{0.081}\\
        \midrule
        \multirow{2}{*}{FB237} &41.8 &12.8 &11.0 &32.6 &47.3 &25.5 &14.0 &14.5 &10.8  &23.4 \\
        &\std{0.058} &\std{0.118} &\std{0.173} &\std{0.084} &\std{0.169} &\std{0.208} &\std{0.153} &\std{0.104} &\std{0.203} &\std{0.050}\\
        \midrule
         \multirow{2}{*}{NELL}   &53.1 &16.1 &13.9 &40.0&50.8 &26.3 &17.5 &15.3 &11.3   &27.2 \\
        &\std{0.117} &\std{0.193} &\std{0.260} &\std{0.119} &\std{0.076} &\std{0.175} &\std{0.154} &\std{0.102} &\std{0.193} &\std{0.071}\\
        \bottomrule
    \end{tabular}

\end{table*}

\begin{table}[ht]
    \centering
    \caption{The mean values and standard variances of ConE's MRR results on queries with negation.}
    \label{table:errorbar_neg}
    \vspace{1mm}
    \begin{tabular}{ c  c c c c  c c c c }
        \toprule
         \textbf{Dataset}  &\textbf{2in} &\textbf{3in} &\textbf{inp} &\textbf{pin} &\textbf{pni} &\textbf{AVG}\\
        \midrule
        \multirow{2}{*}{FB15k} 
        &17.9 &18.7 &12.5 &9.8 &15.1 &14.8\\
        &\std{0.158} &\std{0.206} &\std{0.094} &\std{0.428} &\std{0.172} &\std{0.139}\\
        \midrule
        \multirow{2}{*}{FB237}
        &5.4 &8.6 &7.8 &4.0 &3.6 &5.9\\
        &\std{0.075} &\std{0.076} &\std{0.135} &\std{0.078} &\std{0.069} &\std{0.037}\\
        \midrule
        \multirow{2}{*}{NELL} 
        &5.7 &8.1 &10.8 &3.5 &3.9 &6.4\\
        &\std{0.022} &\std{0.129} &\std{0.199} &\std{0.014} &\std{0.088} &\std{0.054}\\
        \bottomrule
    \end{tabular}
    
\end{table}

\subsection{Union using De Morgan's Law and the Real Union.}
When we use De Morgan's law to approximate the union, the resulted cones are always sector-cones, which may be inconsistent with the real union. We conduct experiments to compare the learned embeddings for $\neg (\neg A\land \neg B)$  and $A\lor B$. Specifically, we randomly generate $8000$  pairs of sector-cones $(A_i, B_i)$  and generate embeddings for $\neg (\neg A_i\cap \neg B_i)$  and $A_i\cup B_i$. Then, to measure the overlap between $\neg (\neg A_i\cap \neg B_i)$  and $A_i\cup B_i$ , we calculate the ratio $r_i=|(\neg (\neg A_i\cap\neg B_i))\cap(A_i\cup B_i) |/|(\neg (\neg A_i\cap\neg B_i))\cup(A_i\cup  B_i)|$ , and obtain an average ratio of $r=0.4618$ . The results show a relatively high discrepancy between $\neg (\neg A_i\cap \neg B_i)$  and $A_i\cup B_i$, which again validates the results that ConE with DNF technique can outperform ConE with De Morgan’s law.

\subsection{Modeling the Variability of Answer Sets}
It is possible that an answer set to a query has a large number of entities but small apertures. When it happens, there are two possible cases.
\begin{enumerate}
    \item The semantic variability of the entities in this set is low. That is to say, entities in the set closely locate in a cone with a small aperture.
    \item Some of the entities are outside the cone. The learned cone embeddings of a query may not include all its answer entities, especially for queries in the validation/test sets. This phenomenon also partly accounts for imperfect performance.
\end{enumerate}

We conduct experiments on FB15k to demonstrate that the learned apertures are correlated with the similarity measures over answer sets. The results are shown in Table \ref{table:variety}. In this experiment, suppose that we have an entity set $[[q]]=\{v_1^q, \dots, v_{n_q}^q\}$ that is the answer set to a query $q$,  and its corresponding embeddings $\{\textbf{v}_1^q,\dots,\textbf{v}_{n_q}^q\}$ (note that their apertures are zero). First, we compute the average embeddings of $\{\textbf{v}_1^q,\dots,\textbf{v}_{n_q}^q\}$ using SemanticAverage (all weights are set to be equal) introduced in Section 5.2. Then, we calculate the maximum squared distance from the entity embeddings to the average embeddings and let $\delta_q$ denote the result. That is, $\delta_q$ measures the overall variation of entities in $[[q]]$. Finally, we calculate the Spearman’s rank correlation and Pearson’s correlation between $\delta_q$ and the learned apertures of $q$.

\begin{table*}[ht]
    \centering
    \caption{Spearman's rank correlation (SRC) and Pearson's correlation (PC) between learned aperture embeddings and the variety of answer sets on FB15k.}
    \label{table:variety}
    \vspace{1mm}
    \setlength{\tabcolsep}{2.0mm}{
    \begin{tabular}{c c c  c c c c  c c c c c c c}
        \toprule
         &\textbf{1p} &\textbf{2p} &\textbf{3p} &\textbf{2i} &\textbf{3i} &\textbf{pi} &\textbf{ip} &\textbf{2in} &\textbf{3in} &\textbf{inp} &\textbf{pin} &\textbf{pni} \\
        \midrule
         SRC &0.28	&0.27	&0.22	&0.58	&0.43	&0.44	&0.27	&0.41	&0.42	&0.10	&0.16	&0.39\\
        PC & 	0.40	&0.28	&0.25	&0.54	&0.49	&0.42	&0.29	&0.33	&0.38	&0.09	&0.11	&0.30\\
        \bottomrule
    \end{tabular}
    }
    \vspace{-1mm}
\end{table*}

\section{Semantic Average and Ordinary Average}
We give a figure illustration of the difference between the ordinary and semantic average. When $d=1$, if $\gb{\theta}_{1,\taxis}=\pi-\epsilon$ and $\gb{\theta}_{2,\taxis}=-\pi+\epsilon$ ($0<\epsilon<\pi/4$), then we expect $\gb{\theta}_{\taxis}$ to be around $\pi$. However, if we use the ordinary weighted sum, $\gb{\theta}_{\taxis}$ will be around $0$ with a high probability.

\begin{figure}[!ht]
    \centering
    \includegraphics[width=0.6\columnwidth]{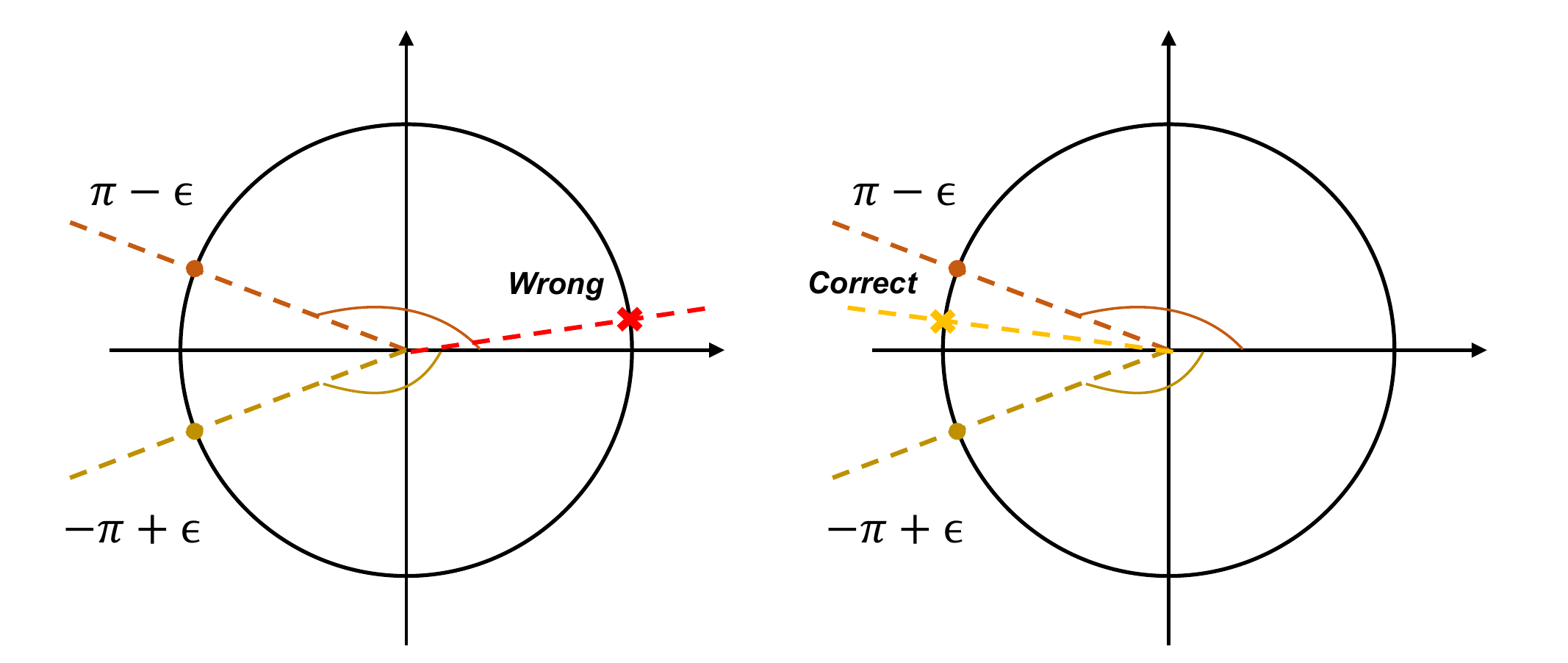}
    \caption{Illustrations of two different weighted average protocols. The left figure represents the ordinary weighted average, which leads to a wrong result. The right figure represents the proposed semantic average.}
    \label{fig:intersect}
\end{figure}

\section{Determining Whether an Entity Belonging to an Answer Set}
Whether an entity $v$ belongs to the answer set $[[q]]$ is determined by the outside distance $d_o(\textbf{V},\textbf{V}_q)$, where $\textbf{V}$ and $\textbf{V}_q$ are the cone embeddings for the entity $v$ and query $q$, respectively. Ideally, the entity $v$ belongs to the answer set $[[q]]$  when $d_o(\textbf{V},\textbf{V}_q)=0$, i.e., the entity embedding $\textbf{V}$ intersects all of the cones in the query embedding $\textbf{V}_q$. Accordingly, in the ideal case, an entity belongs to the complement if it intersects all of the cones in the negation query embedding. However, we allow some components of $\textbf{V}$ outside the corresponding components of $\textbf{V}_q$ in practice. If $d_o(\textbf{V},\textbf{V}_q)$ is small enough (e.g., smaller than a threshold), we can recognize the entity $v$ as an answer to the query $q$. Moreover, in this way, even if we have two entities that both have mismatched cones, we can say that one entity is more likely to be the answer than the other one by comparing their distances to the query embeddings.

We claim that geometry-based models can determine an entity as an answer to a given query if the cones/boxes represented the entity are inside the cones/boxes represented the query. We conduct experiments to validate the above claim.
Specifically, we use trained models ConE/Query2Box with embedding dimensions $d=800$. That is, each entity and query is represented by a Cartesian product of $800$ cones/boxes. Given an entity embedding $\textbf{v}$ and a query embedding $\textbf{V}_q$, if a majority (we use a threshold of $500$ in the experiments) of the $800$ cones/boxes of $\textbf{v}$ are inside the cones/boxes of $\textbf{V}_q$, we regard the entity $v$ as an answer to the query $q$.  Given a query in the validation/test set, we see its answer entities as positive samples and all the other entities in the KG as negative samples. 

Table \ref{tab:answer_epfo} shows the precision/recall results of the validation/test queries. Note that Query2Box does not apply to queries with negation, so we do not include the corresponding results. The results demonstrate that, using geometry-based models, we can determine whether an entity is an answer to a query by the inclusion relation between entity embeddings and query embeddings. Moreover, ConE outperforms Query2Box on the queries without negation, which is consistent with the results in Table 1 in the main text.
\begin{table}[ht]
    \centering
    \caption{Precision/recall results of determining answer entities on queries without negation. The first two rows are results for validation queries, and the last two rows are results for test queries.}
    \label{tab:answer_epfo}
    \vspace{1mm}
    \begin{tabular}{cccc}
    \toprule
        & \textbf{FB15k} &\textbf{FB237} & \textbf{NELL}\\
        \midrule
        \textbf{Q2B} & 0.490/0.532 &0.483/0.533 &0.458/0.581 \\
        \textbf{ConE} & 	0.580/0.678	&0.519/0.645 &0.583/0.696\\
        \midrule
        \textbf{Q2B} & 0.502/0.527	&0.489/0.506	&0.466/0.562 \\
        \textbf{ConE} &0.610/0.670	&0.536/0.670	&0.604/0.645\\
    \bottomrule
    \end{tabular}
\end{table}

\begin{table}[ht]
    \centering
    \caption{Precision/recall results of determining answer entities on queries with negation. The first row is results for validation queries, and the last row is results for test queries.}
    \label{tab:answer_neg}
    \vspace{1mm}
    \begin{tabular}{cccc}
    \toprule
        & \textbf{FB15k} &\textbf{FB237} & \textbf{NELL}\\
        \midrule
        \multirow{2}{*}{\textbf{ConE}} & 	0.455/0.648	&0.545/0.636	&0.513/0.693\\
        &0.510/0.656	&0.560/0.608	&0.524/0.627\\
    \bottomrule
    \end{tabular}
\end{table}

\newpage
\section{Qualitative Analysis Between ConE and Query2Box}
The embedding space and the operators are two key parts of a query embedding model. Therefore, we introduce the superiority of ConE over Query2Box in these two aspects.

\subsection{Embedding Space}
Cones can naturally represent a finite universal set and its subset, while Query2Box cannot. The universal set in a knowledge graph corresponds to the set consisting of all the entities, which is finite. As the apertures of cones are bounded (between $0$ and $2\pi$), we can use the cones with apertures $2\pi$ to represent the universal set and find cones with proper apertures to represent any subsets of the universal set. However, since the offsets of boxes in Query2Box are unbounded, how to find boxes to represent the universal set is unclear. It is worth noting that we cannot constrain the offsets of boxes in Query2Box to be bounded, since the composition of its projection operator can generate boxes with arbitrarily large offsets.

The axes of cones are periodic while the centers of boxes are not. It is an important property to model symmetric relations. We will discuss it in detail in the next part.

\subsection{Operators}
The operators in query embedding models usually contain projection, intersection, union, and complement.  The superiority of ConE over Query2Box mainly comes from the projection and complement operator.

The projection operator of ConE can generate cones with larger or smaller apertures depending on the relation. However, the projection operator of Query2Box always generates a larger box with a translated center, no matter what the relation is.  In fact, not all the relation projections should result in larger boxes. For example, if an entity set contains all the countries in the world, and the relation is \textit{contain\_cities}, the set of adjacent entities will be larger. If the given entity set contains all cities in the world and the relation is \textit{locate\_in\_country}, the set of adjacent entities will be smaller. Therefore, the projection operator of ConE is more expressive than that of Query2Box. An expressive projection operator can improve the performance on all the queries as projection appears in all query structures. 

The projection operator of ConE can well deal with symmetric relations, while the translation-based projection operator of Query2Box cannot.  Suppose that $r$ is a symmetric relation. That is, if $r(h,t)$  is true, then $r(t,h)$    will also be true (e.g., \textit{married\_with}).  Suppose that the embedding dimension $d=1$   , the axis of $h$ is $\theta_h$, the axis of $t$ is $\theta_t$, and the apertures of $h$ and $t$ are $0$. Then, ConE can model the symmetric relation by learning a neural operator that rotates some axes by an angle $\pi$ and keeps the apertures unchanged. That is, ConE can model the relation $r$ between $h$ and $t$ as $r(\theta_h)=\theta_h+\pi=\theta_t$ and $r(\theta_t)=\theta_t+\pi=\theta_h$, which is benefited from the periodicity. A similar case can be found in RotatE. RotatE can deal with symmetric relations since the phases in complex spaces are periodic.

Since the complements of boxes are no longer boxes, it is still unclear how to use boxes to model the complement operation. 

\section{Computational Complexity}
The computational complexity of ConE is similar to that of Query2Box \cite{q2b}. Given a query in Disjunctive Normal Form $q=q_1\lor \dots\lor q_n$, where $q_i$ are conjunctive queries, the computational complexity of ConE to answer $q$ is equal to that of answering the $n$ conjunctive queries $q_i$. Answering $q_i$ requires to execute a sequence of simple geometric cone operations, each of which takes constant time. Then, we perform a fast search using techniques such as Locality Sensitive Hashing \cite{lsh} to get the final answer.

To evaluate the training speed of ConE and all the baselines, we report the average time spent to run 100 training steps. We run all the models with the same number of embedding parameters using a single RTX 3090 GPU card. Table \ref{tab:running_time} demonstrates that the simplest model GQE is the most time-efficient. The training speed of ConE is close to that of Query2Box (Q2B) and faster than BetaE. 

\begin{table}[ht]
    \centering
    \caption{Average time spent per 100 training steps.}
    \label{tab:running_time}
    \vspace{1mm}
    \begin{tabular}{ccccc}
    \toprule
    \textbf{Models} & \textbf{GQE} & \textbf{Q2B} & \textbf{\betae} & \textbf{ConE}\\
    \midrule
     \textbf{Running Time} & 15s  & 17s  & 28s & 21s\\
     \bottomrule
    \end{tabular}
\end{table}

\section{Potential Societal Impacts}
ConE is a method that performs automatic reasoning  over knowledge graphs. One potential negative societal impacts when using automatic reasoning methods (including ConE) is privacy disclosure. If we use public data on the Internet or somewhere else to construct a knowledge graph, and then perform multi-hop reasoning over it, personal information that one does not want to make public may be exposed.
\end{document}